\PassOptionsToPackage{table,xcdraw}{xcolor}
\documentclass[sigconf, nonacm]{acmart}

\renewcommand\footnotetextcopyrightpermission[1]{} % removes footnote with conference information in first column
\usepackage{latexsym}
\usepackage{enumitem}

\usepackage{enumerate}
\usepackage{multirow}
\usepackage{amsmath}
\usepackage{hyperref}

\usepackage{tabularx}
\usepackage{float}

\usepackage[table]{xcolor}
\usepackage{graphicx}
\usepackage{svg}
\usepackage{subfig}
\usepackage{array}
\usepackage{enumitem}

\usepackage{booktabs}
\usepackage{mathtools} %AK
\usepackage{adjustbox} %SAH
\usepackage{stfloats} % ash
\usepackage{siunitx} % ash

\begin{document}

% \title[Finding Already Debunked Narratives via Multistage Retrieval]{Finding Already Debunked Narratives via Multistage Retrieval: Enabling Cross-Lingual, Cross-Dataset and Zero-Shot Learning}
\title{Breaking Language Barriers with MMTweets: Advancing Cross-Lingual Debunked Narrative Retrieval for Fact-Checking}
\pagestyle{plain}


% \affiliation{%

\author{Iknoor Singh, Carolina Scarton, Xingyi Song, Kalina Bontcheva}
\affiliation{%
  \institution{University of Sheffield}
  \city{Sheffield}
  \country{United Kingdom}}
\email{{ i.singh, c.scarton, x.song, k.bontcheva } @sheffield.ac.uk}

% \author{Anonymous Submission}

% \affiliation{%

% \affiliation{%

% \renewcommand{\shortauthors}{Anonymous.}

\begin{abstract}
Finding previously debunked narratives involves identifying claims that have already undergone fact-checking. The issue intensifies when similar false claims persist in multiple languages, despite the availability of debunks for several months in another language. Hence, automatically finding debunks (or fact-checks) in multiple languages is crucial to make the best use of scarce fact-checkers' resources. Mainly due to the lack of readily available data, this is an understudied problem, particularly when considering the cross-lingual scenario, i.e. the retrieval of debunks in a language different from the language of the online post being checked. This study introduces cross-lingual debunked narrative retrieval and addresses this research gap by: (i) creating Multilingual Misinformation Tweets (MMTweets): a dataset that stands out, featuring cross-lingual pairs, images, human annotations, and fine-grained labels, making it a comprehensive resource compared to its counterparts; 
(ii) conducting an extensive experiment to benchmark state-of-the-art cross-lingual retrieval models and introducing multistage retrieval methods tailored for the task; and (iii) comprehensively evaluating retrieval models for their cross-lingual and cross-dataset transfer capabilities within MMTweets, and conducting a retrieval latency analysis. We find that MMTweets presents challenges for cross-lingual debunked narrative retrieval, highlighting areas for improvement in retrieval models. Nonetheless, the study provides valuable insights for creating MMTweets datasets and optimising debunked narrative retrieval models to empower fact-checking endeavours. The dataset and annotation codebook are publicly available at \url{https://doi.org/10.5281/zenodo.10637161}.
\end{abstract}


%    <concept>
%        <concept_id>10002951.10003227.10003351</concept_id>
%        <concept_desc>Information systems~Data mining</concept_desc>
%        <concept_significance>500</concept_significance>
%        </concept>
%    <concept>
%        <concept_id>10002951.10003260.10003282.10003292</concept_id>
%        <concept_desc>Information systems~Social networks</concept_desc>
%        <concept_significance>500</concept_significance>
%        </concept>
%    <concept>
%        <concept_id>10010147.10010178.10010179.10003352</concept_id>
%        <concept_desc>Computing methodologies~Information extraction</concept_desc>
%        <concept_significance>500</concept_significance>
%        </concept>
%  </ccs2012>

% \ccsdesc[500]{Information systems~Data mining}
% \ccsdesc[500]{Information systems~Social networks}
% \ccsdesc[500]{Computing methodologies~Information extraction}



\maketitle

\section{Introduction}

\begin{figure}[]
    \centering
\scalebox{0.9}{    \includegraphics[width=9.4cm,height=3cm]{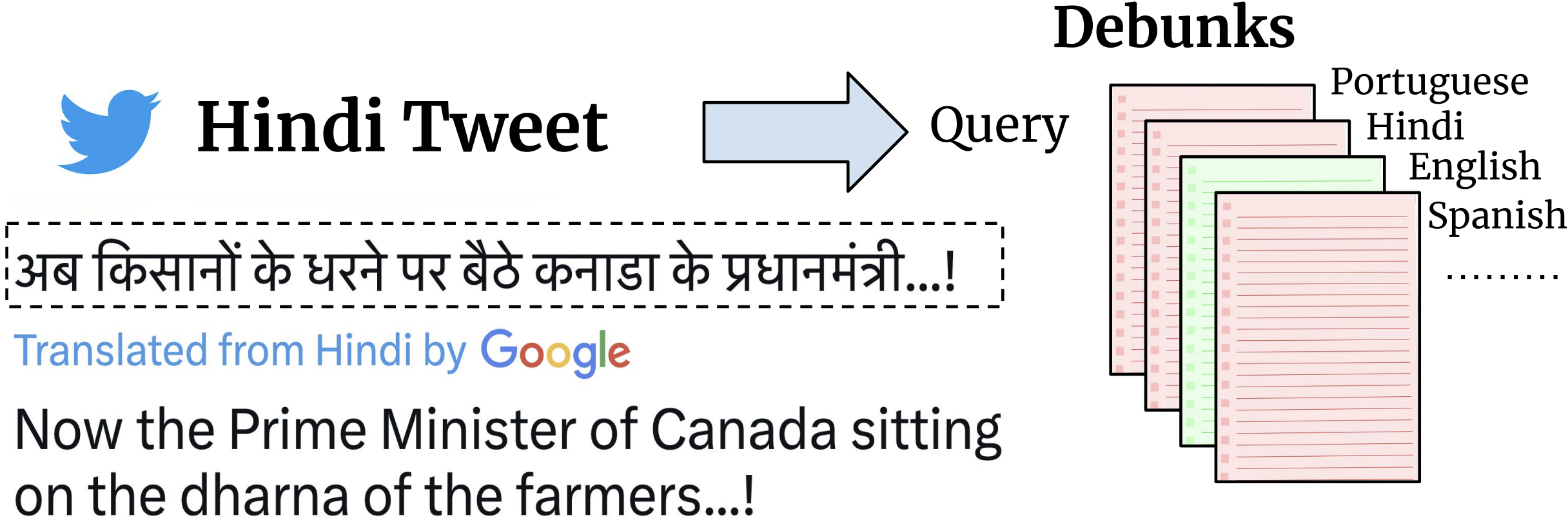}
}
    % \vspace{-5pt}
    % \caption{X-DNR: Hindi tweet and an English relevant debunk .}
\caption{Cross-lingual debunked narrative retrieval: Query tweet is in Hindi and the relevant debunk is in English.}

    % \vspace{-10pt}
    \label{fig:pipe}
\end{figure}

% \begin{figure*}[]
%     \centering
%    \includegraphics[width=\textwidth]{chart (1).png}
%     % \vspace{-5pt}
%     \caption{Cross-lingual retrieval of already debunked narratives: tweet is in Hindi and the relevant debunk is in English.}
%     % \caption{Cross-lingual debunked narrative retrieval: Hindi tweet and English relevant debunk.}

%     % \vspace{-10pt}
%     \label{fig:pipe}
% \end{figure*}

Automated fact-checking systems play a vital role in both countering false information on digital media and alleviating the burden on fact-checkers \cite{nielsen2022mumin, nielsen2022mumin, shang2023amica, wu2022bias, guo2022survey, zhang2022learning, zeng2021automated}. A key task of these systems is the detection of previously fact-checked similar claims -- an information retrieval problem where claims serve as queries to retrieve from a corpus of debunks \citep{nakov2022overview, nakov2021clef, shaar2020overview}. This task aims to detect claims that spread even after they have already been debunked by at least one professional fact-checker. Previous work has focused on training retrieval models, primarily focusing on monolingual retrieval, where the language of the query claim matches the language of the debunk  \cite{nakov2022overview, nakov2021clef, shaar2020overview, kazemi2021claim}. Moreover, these monolingual retrieval models assume that the debunks exist exclusively in one language. However, previous studies \citep{singh2022comparative, singh2021false, reis2020can} demonstrate that similar false claims continue to spread in multiple languages, despite the availability of debunks for several months in another language. Hence, automatically finding debunks in multiple languages is crucial to make the best use of scarce fact-checkers' resources. 

% https://semakanfakta.afp.com/pelbagai-video-menunjukkan-serangan-lebah-di-nicaragua-dan-mangsa-kebocoran-gas-loji-kimia-di-india
\begin{table*}[!t]
\small
\centering
\caption{
Sample query tweets and their corresponding debunks from the \textbf{MMTweets} dataset.
}
\adjustbox{max width=\textwidth}{
\begin{tabular}{p{2.5cm}p{5.5cm}p{7cm}} 
\toprule
\textbf{Fields} & \textbf{Hindi Query Tweet - English Debunk}  & \textbf{English Query Tweet - Spanish Debunk}                             \\ \midrule
\textbf{Tweet} & \raisebox{-0.70\totalheight}{\includegraphics[width=5.2cm, height=1cm]{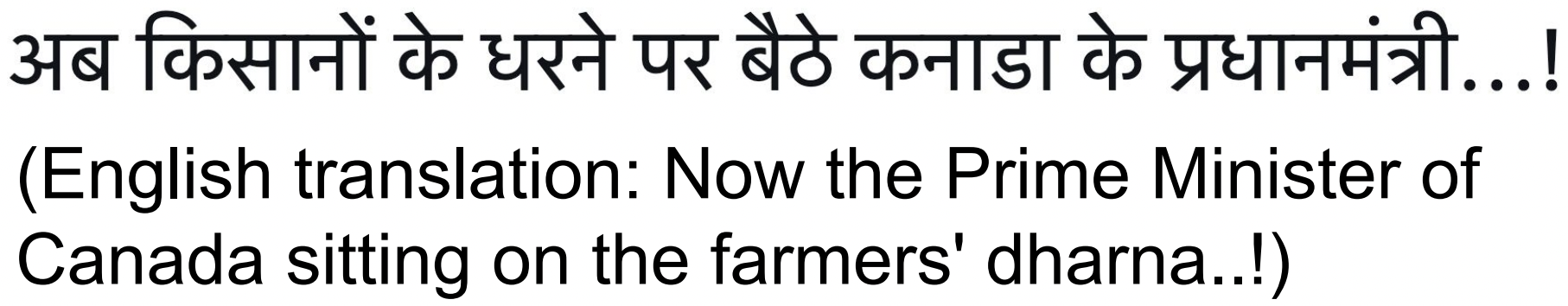}} & I Sultue you Sir. You are So intelligent. RUSSIA: Vladimir Putin has Dropped 800 tigers and Lions all over the Country to push people to stay Home...Stay Safe Everyone!!                \\ \midrule
\textbf{Debunk title} & Old Photo Passed Off As Justin Trudeau Sitting In An Anti-Farm Laws Protest & La foto del león en la calle fue tomada en Sudáfrica en 2016 y no tiene relación con la pandemia del COVID-19 \\ \midrule
\textbf{Debunk claim} & Justin Trudeau sits in protest in support of the protesting farmers.  & Publicaciones compartidas más de 35.000 veces en redes sociales desde el 22 de marzo último aseguran que Rusia liberó..  \\
% \textbf{Debunk article} & A photo from 2015 of Canadian Prime Minisiter Justin Trudeau attending a Diwali celebration  [...] & Current technologies and international legal and health controls would not allow the introduction of a chip in future [...] \\ 
\bottomrule
\end{tabular}}
% \vspace{-5pt}
% \vspace{-15pt}
% \caption{
% Sample Hindi and Spanish query tweets and their corresponding debunks from the \textbf{MMTweets} dataset.
% }
\label{tab:example}
\end{table*}

In this study, we define the task of \textbf{cross-lingual Debunked Narrative Retrieval (X-DNR)} as a cross-lingual information retrieval problem where a claim is used as a query to retrieve from a corpus of debunks in multiple languages (see Figure \ref{fig:pipe}). In this paper, we use the term ``debunked narrative retrieval'' over the previously used term ``fact-checked claim retrieval'' because the term ``debunked narrative'' better captures the range of false narratives or stories related to a claim that has already been debunked. This term acknowledges that a single claim can have multiple narratives, all needing debunking, unlike fact-checked claim retrieval, which focuses narrowly on verified claims without addressing their diverse associated narratives. Therefore, the term ``debunked narrative retrieval'' is more fitting for this task, as the primary objective of X-DNR is to aid fact-checkers in identifying debunked narratives across multiple languages.
% The main contributions of this paper are:
Our main contributions are:

\begin{itemize}[noitemsep, leftmargin=*]

\item The \textbf{M}ultilingual \textbf{M}isinformation \textbf{Tweets} (\textbf{MMTweets}): a novel benchmark that stands out, featuring cross-lingual pairs, images, and fine-grained human annotations, making it a comprehensive resource compared to its counterparts (see Section \ref{comparisondata}). In total, it comprises $1,600$ query tweet claims (in Hindi, English, Portuguese \& Spanish) and $30,452$ debunk corpus (in 11 different languages) for retrieval. Table \ref{tab:example} shows dataset examples.
% \footnote{The dataset and code are available at https://doi.org/10.5281/zenodo.7144808.}
% https://doi.org/10.5281/zenodo.7144808
% (examples in Table \ref{tab:example}).

\item An extensive evaluation of state-of-the-art (SOTA) cross-lingual retrieval models on the MMTweets dataset. We also introduce two multistage retrieval methods (\textit{BE+CE} and \textit{BE+GPT3.5}) adapting earlier approaches to effectively address the cross-lingual nature of the X-DNR task. Nevertheless, the results suggest that dealing with multiple languages in the MMTweets dataset poses a challenge, and there is still room for improvement in models.

\item A comprehensive evaluation aims to investigate: 1) cross-lingual transfer and generalisation across languages within MMTweets; 2) how challenging it is for models trained on existing datasets to transfer knowledge to the MMTweets test set; 
% 3) the impact of the type and count of negative pairs on the model's performance;
and 3) insights into the retrieval latency of different models (see Section \ref{resandiscuss}).
% A comprehensive evaluation of retrieval models trained on MM-Tweets investigates challenges in cross-lingual and cross-dataset transfer, sensitivity to negative claim and debunk pairs, and offers insights into the retrieval latency of different models.

\end{itemize}

In the following section, we discuss the related work. Section \ref{mmtweetsdataset} details the MMTweets dataset. Section \ref{debunkSearchMethods} presents the various experimental details related to the X-DNR task. The results are presented in Section \ref{resandiscuss} and we conclude the paper in Section \ref{debunksearchconclusion}.

\section{Related Work}
\label{debunksearchrelatedwork}

% \subsection{Cross-lingual Retrieval of Already Debunked Narratives}
In order to minimise the spread of misinformation and speed up professional fact-checking, the initial verification step often involves searching for fact-checking articles that have already debunked similar narratives \citep{singh2023utdrm, la2023retrieving, shaar2021role, shaar-etal-2020-known}. Several benchmark datasets have been created for this task \citep{nakov2022overview, nakov2021clef, shaar2021assisting, mansour2023not, mansour2022did, sheng2021article, chakraborty2023information}.
% This task is accomplished by using retrieval models\cite{singh2023utdrm, la2023retrieving, shaar2021role, shaar-etal-2020-known}, which use misinformation claims as queries to find relevant debunked narratives \cite{mansour2023not, mansour2022did, sheng2021article, chakraborty2023information}. 
For instance, \citet{shaar-etal-2020-known} release a dataset of English claims and fact-checking articles from Snopes\citep{snopesSnopescom} and PolitiFact \citep{politifactPolitiFact}. On the other hand, \citet{vo2020facts} release a multimodal English dataset of tweet claims collected from Snopes and PolitiFact and investigates the use of images in tweets to retrieve previously fact-checked content. The CLEF CheckThat! Lab evaluations \citep{shaar2020overview, nakov2021clef, nakov2022overview, barron2023clef} focus on a fully automated pipeline of fact-checking claims, where fact-checked claim retrieval is one of the steps in the claim verification workflow. They release a dataset of claims collected from Snopes, PolitiFact and AraFacts \citep{ali2021arafacts} and ClaimsKG \citep{tchechmedjiev2019claimskg}. However, the aforementioned work only focuses on monolingual scenarios where the claim and debunk share the same language. In contrast, our MMTweets dataset includes cross-lingual cases, making it more challenging. For a detailed comparison of different datasets with our MMTweets, please refer to Section \ref{comparisondata}. We also test domain overlap between MMTweets and other datasets in Section \ref{CLRADNresult3}.

Prior work on claim matching \citep{kazemi2021claim} release a dataset of claims collected from tiplines on WhatsApp \citep{kazemi2021claim} and conduct retrieval experiments. Although they present results for multiple languages, their dataset only includes monolingual pairs \citep{kazemi2021claim}, thereby hindering the development of retrieval models capable of detecting debunked narratives in multiple languages. 
Finally, the closest match to our work \citep{kazemi2022matching} focuses on cross-lingual claim matching. They release a dataset of debunked tweets sourced from the International Fact-Checking Network (IFCN)
% \footnote{\url{https://www.poynter.org/ifcn/}} 
\citep{ifcnifcn} and some other fact-checking aggregators \citep{kazemi2022matching}. However, their dataset lacks diverse cross-lingual pairs (see Section \ref{comparisondata}), and tweet claims are automatically extracted from debunk articles \citep{kazemi2022matching}, which can result in false positives. In contrast, our dataset has diverse cross-lingual pairs, and each tweet in MMTweets undergoes manual annotation to ensure high-quality data (see Section \ref{mmtweetsdataset}). Moreover, prior work \citep{kazemi2022matching} does not train custom debunked narrative retrieval models or perform cross-lingual and cross-dataset transfer testing, a gap that we address in this paper with a specific focus on the MMTweets dataset (Section \ref{resandiscuss}). 

 %Next, \citet{kazemi2022matching} focus on matching tweet and debunk pairs, achieving an accuracy of up to 89\% across four language pairs. They find that BM25 outperforms or matches MPT models in monolingual retrieval, while off-the-shelf LaBSE model excels in cross-lingual retrieval.

Furthermore, \citet{kazemi2021claim} found that multistage retrieval \citep{nogueira2019passage} using BM25 and XLM-RoBERTa transformer \citep{conneau2019unsupervised} re-ranking can beat the competitive BM25 baseline for debunked narrative retrieval. However, the use of multistage retrieval with BM25 and transformer model re-ranking, as demonstrated in prior work \citep{kazemi2021claim, shaar-etal-2020-known, nogueira2019passage, thakur2021beir}, introduces translation overhead for BM25 in cross-lingual scenarios where the query claim and document languages differ. To address this, this paper introduces translation-free multistage retrieval methods, employing both bi-encoders and cross-encoders for the X-DNR task (Section \ref{MultistageRetrieval}). Additionally, due to dataset limitation, much of the prior research \citep{shaar-etal-2020-known, nakov2021clef, nakov2022overview} trains retrieval models using debunks available from a single fact-checking organisation. In contrast, our MMTweets dataset involves debunks from multiple fact-checking organisations (Section \ref{mmtweetsdataset}). This enables the development of retrieval models that are agnostic to debunk structure, a crucial aspect for X-DNR, as relevant debunks can originate from any fact-checking organisation.

\section{MMTweets Dataset}
\label{mmtweetsdataset}

MMTweets is a new dataset of misinformation tweets annotated with their corresponding debunks (or fact-checks), both available in multiple languages. MMTweets primarily comprises of tweets related to COVID-19 misinformation in English, Hindi, Portuguese and Spanish. The languages of tweets were selected based on two criteria: 1) these are the most frequent languages in previous publicly available COVID-19 misinformation datasets \citep{li2020mm, singh2021false}; 2) the chosen languages are among some of the most widely spoken ones worldwide. The dataset was built in two steps: first, the raw data was collected, followed by manual data annotation.

\subsection{Raw Data Collection} 
\label{MM Data Colletion}

First, we collect debunk narratives published by different fact-checking organisations covering our target languages. For this, we collect a total of $30,452$ debunk articles from the following organisations (language in brackets): Boomlive
% \footnote{\url{https://www.boomlive.in/}} 
(English) \citep{boomlive}, Agence France-Presse (AFP)
% \footnote{\url{https://www.afp.com/}} 
(German, English, Arabic, French, Spanish, Portuguese, Indonesian, Catalan, Polish, Slovak and Czech) \citep{afpafp}, Agencia EFE
% \footnote{\url{https://www.efe.com/}} 
(Spanish) \citep{efecom} and Politifact
% \footnote{\url{https://www.politifact.com/}} 
(English) \citep{politifactPolitiFact}. For each debunk article, we collect the following information fields: the article title, the debunked claim statement and the article body. 
% Next, we select a sample of $1,600$ debunk articles with a focus on COVID-19 misinformation published between January 2020 and March 2021, so as to allow for temporal and topical diversity as the pandemic unfolded. 

Next, we select a sample of $1,600$ debunk articles from the corpus of $30,452$ debunk articles based on two specific criteria. Firstly, we focus on debunks published between January 2020 and March 2021, allowing for temporal and topical diversity as the COVID-19 pandemic unfolded. This approach, given the global nature of the pandemic, maximises the chance of including similar narratives spreading in multiple languages. Secondly, our aim is to maximise instances where the language of the potential misinformation tweets mentioned in the debunk articles differs from that of the debunk article itself. For example, while Boomlive publishes debunk articles in English, the associated tweets may be in Hindi. Overall, this careful selection of debunks ensures comprehensive cross-lingual coverage within the MMTweets dataset (Section \ref{Linguisticdiv}).

Finally, following the previous work \citep{shaar-etal-2020-known, kazemi2022matching}, we extract all the tweets mentioned in the debunk article body. We use Twitter API
% \footnote{\url{https://developer.twitter.com/en/docs/twitter-api}} 
 \citep{tweetapi} to get tweet details including tweet text and attached media (if any). We chose Twitter because of its easy open access as compared to other social media platforms at the time of this study.

% \begin{table}[!htbp]
% \centering
% \small
% % \caption{MMTweets dataset fields}
% \begin{tabular}{lp{4cm}}
% \toprule
% \textbf{Fields} & \textbf{Explanation}                             \\ \midrule
% \texttt{debunkUrl}       & Fact-checking article URL                                       \\ 
% \texttt{debunkTitle}     & Article title                                     \\ 
% \texttt{debunkClaim}     & Debunk claim statement (by a fact-checker) \\ 
% \texttt{debunkArticle}   & Article body                              \\ 
% \texttt{publishedDate}   & Date of publish                                  \\ 
% \texttt{tweetLink}       & Link to tweet                                    \\ 
% \texttt{tweetText}       & Text of tweet                                    \\ 
% \texttt{tweetLang}       & Language of tweet                                \\ \bottomrule
% \end{tabular}
% % \vspace{-5pt}
% % \caption{MMTweets dataset fields}
% % \vspace{-15pt}
% \caption{MMTweets dataset fields}

% \label{tab:datafields}
% \end{table}

\subsection{Data Annotation -- Tweet Classification} 
\label{guidelines}

\begin{table*}[!htbp]
\small
\centering
\caption{Details of the MMTweets dataset: class count, Fleiss Kappa and textual misinformation ratio.
Please note that the class count does not sum up to the total tweet count due to the overlap between textual and non-textual misinformation cases.}
\begin{tabular}{lccccccc}
\toprule
\multirow{2}{*}{\textbf{Language}} & \multirow{2}{*}{\begin{tabular}[c]{@{}c@{}} \textbf{Tweet Count} \end{tabular}} & \multicolumn{4}{c}{\textbf{Class Count}}                                          & \multirow{2}{*}{\begin{tabular}[c]{@{}c@{}} \textbf{Fleiss Kappa} \end{tabular}} & \multirow{2}{*}{\begin{tabular}[c]{@{}c@{}} \textbf{Textual Misinformation} \\ \textbf{Ratio} \end{tabular}} \\ 
%\cline{2-5}
\cmidrule{3-6}
                          & & \multicolumn{1}{p{1.7cm}}{\textbf{Textual Misinformation}} & \multicolumn{1}{p{2.2cm}}{\textbf{Non-textual Misinformation}} & \multicolumn{1}{c}{\textbf{Debunk}} & \textbf{Other} &                                                                          &                           \\ \midrule
\textbf{Hindi}        & 400             & \multicolumn{1}{c}{328}  & \multicolumn{1}{c}{254}          & \multicolumn{1}{c}{11}     & 27    & 0.53                                                                     & 0.86                      \\ 
\textbf{Portuguese}     & 400           & \multicolumn{1}{c}{310}  & \multicolumn{1}{c}{200}          & \multicolumn{1}{c}{5}      & 30    & 0.59                                                                     & 0.77                      \\ 
\textbf{English}       & 400            & \multicolumn{1}{c}{247}  & \multicolumn{1}{c}{166}          & \multicolumn{1}{c}{68}     & 82    & 0.79                                                                     & 0.61                      \\ 
\textbf{Spanish}      & 400             & \multicolumn{1}{c}{291}  & \multicolumn{1}{c}{233}          & \multicolumn{1}{c}{14}     & 62   & 0.57                                                                     & 0.70                      \\ \midrule
\textbf{Total}        & 1600             & \multicolumn{1}{c}{1176} & \multicolumn{1}{c}{853}          & \multicolumn{1}{c}{98}     & 201   &                                        Average: 0.62                                 &          Average: 0.74                 \\ \bottomrule
\end{tabular}
% \vspace{-5pt}
% \caption{Details of the MMTweets dataset: class count, Fleiss Kappa and textual misinformation ratio.
% Please note that the class count does not sum up to the total tweet count due to the overlap between textual and non-textual misinformation cases.}
\label{tab:MMFlies}
\end{table*}

The approach described in Section \ref{MM Data Colletion} does not guarantee that the extracted tweets from debunk articles contain text-based misinformation. We found that some contained only images or videos, while others made general comments or debunked the misinformation itself. Therefore, the extracted tweets were classified manually to create gold-standard data for evaluation. In particular, we recruited 12 student volunteers\footnote{The dataset annotation received ethical approval from the University of Sheffield Ethics Board (Application ID 040156). This paper only discusses analysis results in aggregate, without providing examples or information about individual users.} who were native speakers of either English, Hindi, Portuguese or Spanish (three native speakers per language). The annotators were shown all debunk information fields and asked to annotate the tweets as belonging to one of three classes: 

\begin{itemize}[noitemsep, leftmargin=*]
    \item \textbf{Misinformation tweets:} with two sub-classes -- \textbf{A) Textual misinformation}, if the textual part of a tweet expresses the false claim which is being debunked by the fact-checking article. 
    \textbf{B) Non-textual misinformation}, if a tweet contains misinformation in image or video only. Please note that a tweet can have both text and non-textual misinformation. For such cases, annotators were asked to label the tweet as having both “textual misinformation” and “non-textual misinformation”.
    \item \textbf{Debunk tweets:} If the tweet does not express misinformation uncritically, but instead exposes the falsehood of the claim. 
    \item \textbf{Other tweets:} If the tweet is neither “misinformation” nor “debunk”, then it is classified as “other”. For instance, this can be a general comment or a general enquiry relevant to the false claim that is being debunked.
\end{itemize}

Please refer to the annotation codebook\footnote{\url{https://doi.org/10.5281/zenodo.10637161}} for examples of misinformation, debunk, and other tweets.
To ensure data quality, we first conducted training sessions with the annotators and went through several examples to familiarise them with the task. We also had a final adjudication step, where problems and disagreements flagged by the annotators were resolved by domain experts. For instance, there were some tweets which agreed with the misinformation but did not state it directly or the annotator was unsure about the claim's veracity. All such cases were considered ``other'' due to the chosen narrower definition of misinformation tweets.  

A total of $1,600$ tweets were annotated, resulting in approximately $400$ tweets per language (see Table \ref{tab:MMFlies}). Following previous methodology \citep{sheng2022characterizing, kazemi2022matching}, a total of $400$ tweets ($100$ per language) were triple annotated to compute inter-annotator agreement (IAA) and the final category was chosen by majority voting. Table \ref{tab:MMFlies} reports Fleiss Kappa scores which indicate moderate to substantial IAA for all languages. Table \ref{tab:MMFlies} also shows the textual misinformation ratio (i.e. the proportion of tweets annotated as ``textual misinformation'' out of all annotated tweets) for each language. The ratio is variable due to the varied nature of the debunks in each language and the different ways in which fact-checkers refer to misinformation-bearing tweets. On average, textual misinformation comprised $74 \%$ of all the classified tweets in the dataset.

\subsection{Data Annotation -- Claim Matching}
\label{claimmatchannotate}

The annotations gathered in Section \ref{guidelines} only pertain to tweets mentioned in the debunk articles, indicating a one-to-one relationship between tweets and debunks. However, prior research \citep{singh2021false, singh2022comparative} demonstrates that there can be various potential debunks for the same misinformation. To address this and establish a one-to-many relationship between misinformation tweets and debunks, we conduct a subsequent round of annotations to identify comparable debunks. However, annotating relevance judgments between tweets and all the previously collected 30,452 debunks is not feasible. Therefore, we take debunked claim statements linked to each tweet and compute cosine similarity\footnote{We use the best performing Sentence-transformer model \textit{all-mpnet-base-v2} on English-translated statements (Ref. \url{https://www.sbert.net/docs/pretrained_models.html})} with all 30,452 debunked claim statements in the hope of finding similar debunked claim statements. To ensure this, we select the \textit{top-k} matching claim statements for annotation, with a depth of seven as per previous work \citep{voorhees2021trec}. We also retain only those claim pairs with a similarity score exceeding the 0.6 threshold to exclude irrelevant claim pairs from the annotations. Finally, annotators classified 4,594 pairs of debunked claim statements into exact match, partial match, or irrelevant (3-level) using previously published annotation guidelines \citep{kazemi2021claim}. Examples for each class can be found in the annotation codebook\footnote{\url{https://doi.org/10.5281/zenodo.10637161}}.

The annotations were conducted on the GATE Teamware annotation tool \citep{bontcheva2013gate} -- refer to the annotation codebook for examples of the tool's user interface. A total of 14 PhD researchers were recruited to manually annotate pairs of debunked claim statements. To ensure high-quality annotations, we conducted a pre-annotation phase. An initial annotator training session familiarises them with the instructions. Subsequently, annotators were asked to annotate a certain number of test samples. We then review these annotations and only those annotators who correctly classify at least 80\% of the samples proceed with further annotations. Based on prior research \citep{mu2023vaxxhesitancy}, we also ask annotators to provide a confidence score for each annotation, and we further discard annotations with low confidence scores to maintain data quality. Finally, following prior works \citep{voorhees2021trec, hu2023mr2, bonisoli2023dice}, we find the IAA Kappa to be 0.5 on a subset of the data using triple annotations, suggesting a moderate level of agreement among the annotators. All annotators were paid at a standard rate of 15 GBP per hour for their work.

% https://dl.acm.org/doi/pdf/10.1145/3539618.3591912

Table \ref{tab:completedata} presents a summary of the complete MMTweets dataset,  including the number of query tweets and the count of query tweet and debunk pairs for 3-level relevance annotations.  Specifically, it includes 2,716 exact matches, 1,542 partial matches, and the remaining are categorised as irrelevant (see Section \ref{Linguisticdiv} for count in different language pairs). The average word count in query tweets is $28\pm14.3$ (1 std). There are a total of 1,600 tweets in MMTweets, and on average, each tweet is linked with $2.7 \pm 2.0$ (1 std) debunks, either exact or partial match. Please note that the one-to-many relation between query tweets and the debunks enriches our dataset to include cases beyond the tweets mentioned in the debunk articles. Additionally, the fine-grained classification of debunks into exact and partial matches serves as fine-grained labels for our subsequent information retrieval experiments (see Section \ref{CLRmodels}).

\subsection{Comparison to Existing Datasets} 
\label{comparisondata}

Table \ref{tab:dnrdatasets} provides a comparison between MMTweets and the existing datasets, revealing favourable query claim counts in our dataset compared to others. Notably, MMTweets stands out with 43\% cross-lingual instances across various language pairs (see Section \ref{Linguisticdiv}). This is in stark contrast to the sole existing cross-lingual dataset \citep{kazemi2022matching}, which only comprises 10\% of Hindi-English pairs, where the claim is in Hindi and the debunk is in English. Additionally, all tweets in MMTweets undergo manual annotation, unlike other existing datasets \citep{hardalov2022crowdchecked, kazemi2022matching, shaar-etal-2020-known}, where tweets are automatically extracted from fact-check articles, potentially leading to false positives. Moreover, automated extraction of tweets also leads to missing one-to-many connections between claims and debunks as shown in prior work \citep{singh2023utdrm}. Furthermore, MMTweets provides 3-level graded relevance scores (fine-grained) for query-passage pairs, unlike prior work which use binary relevance scores (coarse-grained) \citep{shaar-etal-2020-known, nakov2022overview, nakov2021clef}.

Among other datasets, \citet{shaar-etal-2020-known} and CLEF variants lack cross-lingual pairs, images, and fine-grained labels. The larger \citet{vo2020facts} dataset incorporates images and human annotations but lacks fine-grained labels. CrowdChecked \citep{hardalov2022crowdchecked} contains a massive volume of claims but lacks crucial features like manual annotations and cross-lingual pairs. Although prior work \citep{kazemi2022matching, kazemi2021claim} provide multilingual support, it's impossible to replicate or conduct comparative experiments on their datasets because they do not release the corpora of debunks used in the retrieval experiments -- only the query claims are released. Moreover, it lacks images, manual annotations and fine-grained labels \citep{kazemi2022matching}. In contrast, our MMTweets dataset stands out, featuring cross-lingual pairs, images, human annotations, and fine-grained labels, making it a comprehensive resource compared to its counterparts. Additionally, we examine the domain overlap between MMTweets and other datasets, revealing a low degree of overlap (refer to Section \ref{CLRADNresult3}).

\begin{table}[]
\centering
\small
\caption{Complete summary of the MMTweets dataset.}
\resizebox{\columnwidth}{!}{%
\begin{tabular}{lccccc}
\toprule
\textbf{Language}      & \textbf{Hindi} & \textbf{Portuguese} & \textbf{English} & \textbf{Spanish} & \textbf{Total} \\ \midrule
\textbf{Query Tweets}  & 400            & 400                 & 400              & 400              & 1600           \\
\textbf{Exact Match}   & 518            & 742                 & 812              & 644              & 2716           \\
\textbf{Partial Match} & 417            & 409                 & 342              & 374              & 1542           \\
\textbf{Irrelevant}    & 475            & 656                 & 337              & 468              & 1936   
\\ \bottomrule
\end{tabular}%
}
\label{tab:completedata}
\end{table}

\begin{table}[]
\small
\caption{Comparison of debunked narrative retrieval datasets: ``Lang'' denotes the count of different languages of claims; ``Cross'' indicates the presence of cross-lingual pairs; ``Img'' indicates whether the dataset is multi-modal and includes images; ``Ant'' indicates whether the dataset is human-annotated or automatically extracted from articles; ``Fine'' indicates the availability of fine-grained labels.}
\begin{tabular}{lllllll}
\toprule
\multicolumn{1}{c}{\textbf{Dataset}} & \multicolumn{1}{l}{\textbf{Items}} & \multicolumn{1}{l}{\textbf{Lang}} & \multicolumn{1}{l}{\textbf{Cross}} & \multicolumn{1}{l}{\textbf{Img}}  & \multicolumn{1}{l}{\textbf{Ant}}   & \multicolumn{1}{l}{\textbf{Fine}}  \\ \midrule

\citet{shaar-etal-2020-known}                              & 1,768                                & 1       &       \textcolor{red}{$\times$}  &       \textcolor{red}{$\times$} &       \textcolor{red}{$\times$}         & \textcolor{red}{$\times$} \\

CLEF20-EN                   & 1,197                               & 1             &       \textcolor{red}{$\times$} &       \textcolor{red}{$\times$} &       \textcolor{blue}{\checkmark}        & \textcolor{red}{$\times$} \\

CLEF21 2A-EN        & 2,070                              & 1          &       \textcolor{red}{$\times$}   &       \textcolor{red}{$\times$} &       \textcolor{red}{$\times$}          & \textcolor{red}{$\times$} \\

CLEF21 2A-AR      & 858                                & 1      &       \textcolor{red}{$\times$}   &       \textcolor{red}{$\times$} &       \textcolor{blue}{\checkmark}              & \textcolor{red}{$\times$} \\

CLEF22 2A-EN        & 2,362                               & 1         &       \textcolor{red}{$\times$}   &       \textcolor{red}{$\times$} &       \textcolor{red}{$\times$}           & \textcolor{red}{$\times$} \\

CLEF22 2A-AR      & 908                                & 1          &       \textcolor{red}{$\times$}   &       \textcolor{red}{$\times$} &       \textcolor{blue}{\checkmark}           & \textcolor{red}{$\times$} \\

\citet{vo2020facts}                              & 13,239                                & 1          &       \textcolor{red}{$\times$}    & \textcolor{blue}{\checkmark} & \textcolor{blue}{\checkmark}        & \textcolor{red}{$\times$} \\

CrowdChecked \citep{hardalov2022crowdchecked}                          & 330,000                               & 1           &       \textcolor{red}{$\times$}  & \textcolor{red}{$\times$} & \textcolor{red}{$\times$}    & \textcolor{red}{$\times$}      \\

\citet{kazemi2021claim}             & 382                              & 5    &       \textcolor{red}{$\times$}  & \textcolor{red}{$\times$} & \textcolor{blue}{\checkmark} & \textcolor{blue}{\checkmark}\\

\citet{kazemi2022matching}                        & 6,533                               & 4       &       \textcolor{blue}{\checkmark}   & \textcolor{red}{$\times$} & \textcolor{red}{$\times$}  & \textcolor{red}{$\times$}   \\ \hline

\textbf{MMTweets (ours)}                             & 1,600                               & 4 & \textcolor{blue}{\checkmark} & \textcolor{blue}{\checkmark} & \textcolor{blue}{\checkmark}  & \textcolor{blue}{\checkmark}  \\ \bottomrule

\end{tabular}
\label{tab:dnrdatasets}
\end{table}

\begin{table*}[!htbp]
\centering
\small
\caption{Language of tweet and debunk pairs in MMTweets. Language codes are ISO 639-1 representations for Portuguese (PT), Spanish (ES), Hindi (HI), English (EN), Indonesian (ID), Slovak (SK), Catalan (CA), Polish (PL), Czech (CS), and French (FR).}
\begin{tabular}{@{}llllllllllllllllllllll@{}}
\toprule
 \textbf{Tweet Language }  & PT   & ES  & HI  & EN  & EN  & EN  & PT & EN & EN & ES & EN & EN & EN & PT & EN & ES & HI & PT & HI & ES & Total\\ 
\textbf{Debunk Language}   & PT   & ES  & EN  & EN  & ES  & ID  & ES & PT & SK & CA & PL & CA & CS & ID & FR & ID & PT & EN & FR & EN & \\ \midrule
\textbf{Count}   & 1045 & 954 & 925 & 450 & 332 & 158 & 80 & 65 & 53 & 50 & 30 & 27 & 22 & 22 & 17 & 11 & 7  & 4  & 3  & 3  & 4,258\\ \bottomrule
\end{tabular}
% \caption{MMTweets language diversity.}
\label{tab:mmtweets_langdist}
\end{table*}

% \begin{figure*}[]
%     \centering
% \scalebox{0.6}{    \includegraphics[width=19cm,height=10cm]{wordcloud.png}
% }
% % \scalebox{0.6}{    \includegraphics[width=29cm,height=8cm]{lda.png}
% % }
%     % \vspace{-5pt}
%     \caption{Wordcloud for English-translated Hindi (top-left), Portuguese (top-right), English (bottom-left) and Spanish (bottom-right) tweets in the MMTweets dataset.}
%     % \caption{Cross-lingual retrieval of already debunked narratives (CLRADN): Hindi tweet and English relevant debunk.}

%     % \vspace{-10pt}
%     \label{fig:MMWords}
% \end{figure*}

\begin{figure}[]
    \centering
\scalebox{0.9}{    \includegraphics[width=6cm,height=5.6cm]{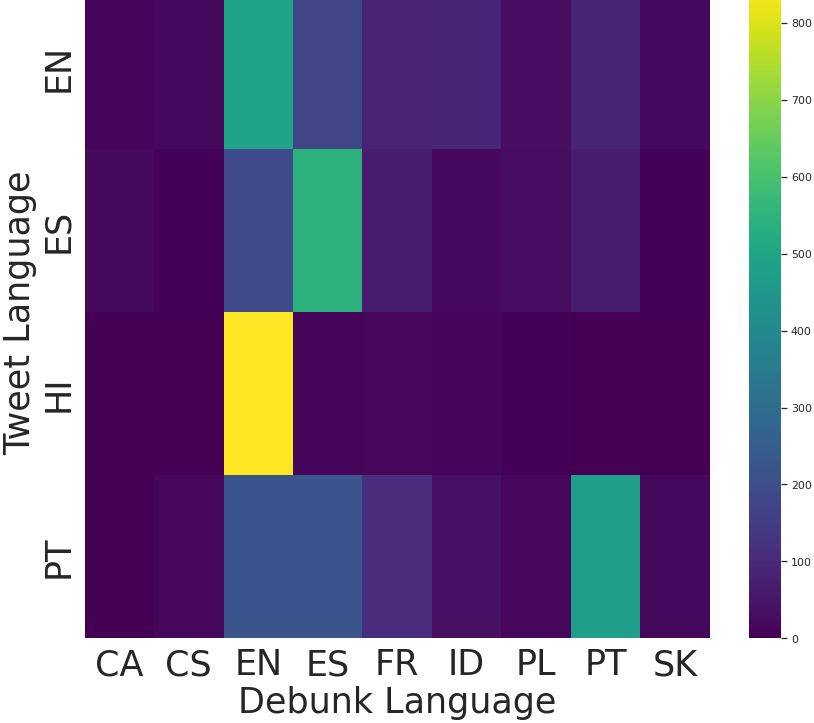}
}
    % \vspace{-5pt}
    \caption{Cross-language analysis: tweet vs. debunk.}
    % \caption{Cross-lingual retrieval of already debunked narratives (CLRADN): Hindi tweet and English relevant debunk.}

    % \vspace{-10pt}
    \label{fig:heatmap}
\end{figure}

\begin{figure}[]
    \centering
\scalebox{0.9}{    \includegraphics[width=8.5cm,height=4.2cm]{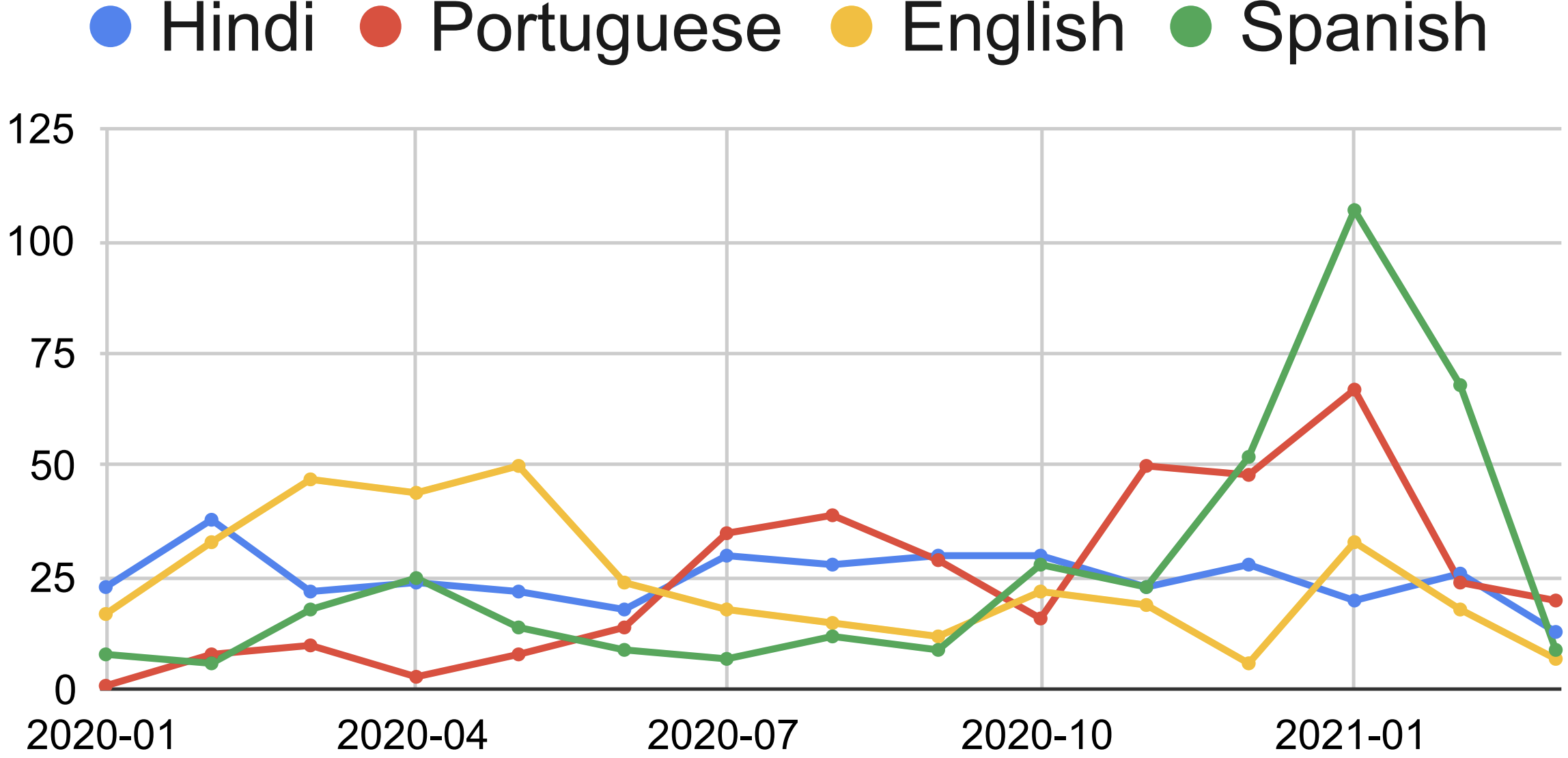}
}
    % \vspace{-5pt}
    \caption{Line plot for month-by-month breakdown of tweet counts for each language in the MMTweets dataset.}
    % \caption{Cross-lingual retrieval of already debunked narratives (CLRADN): Hindi tweet and English relevant debunk.}

    % \vspace{-10pt}
    \label{fig:MMDate}
\end{figure}

\begin{figure}[]
    \centering
\scalebox{0.9}{    \includegraphics[width=7cm,height=3.9cm]{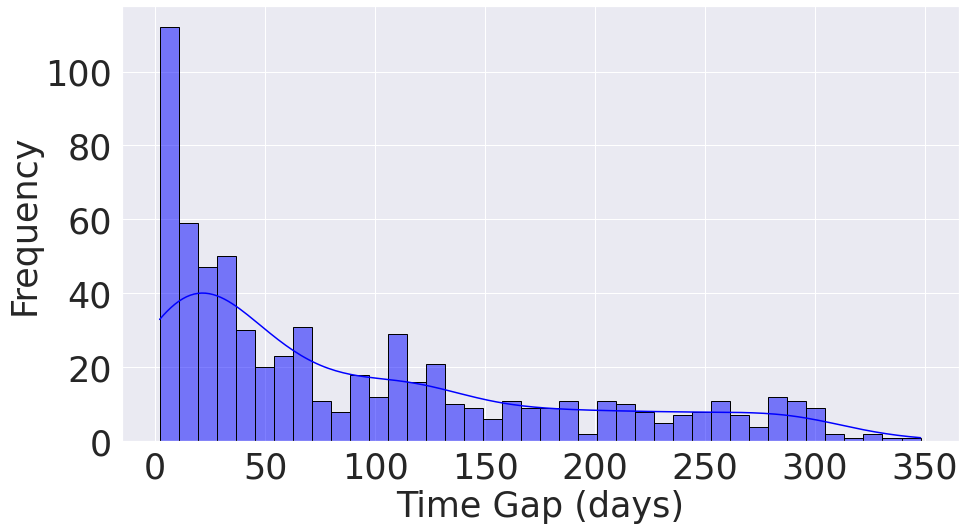}
}
    % \vspace{-5pt}
    \caption{Time gap between tweet and debunk.}
    % \caption{Cross-lingual retrieval of already debunked narratives (CLRADN): Hindi tweet and English relevant debunk.}

    % \vspace{-10pt}
    \label{fig:gaps}
\end{figure}

\subsection{Data Analysis} 
\label{statistics}

\subsubsection{Linguistic Diversity}
\label{Linguisticdiv}
Table \ref{tab:mmtweets_langdist} shows the count of query tweet and debunk pairs for different languages\footnote{We use $langdetect$  (\url{https://pypi.org/project/langdetect/}) for detecting the language.}. In particular, there are a total of 4,258 positive pairs (exact and partial matches) of tweets and their corresponding debunks. Among these, 1,809 instances (43\%) are pairs where the language of tweets and debunks is different (cross-lingual). This makes our dataset the one with the highest proportion of cross-lingual instances when compared to existing datasets (see Section \ref{comparisondata}). The majority of these cross-lingual pairs have tweets in Hindi and corresponding debunks in English, followed by instances with tweets in English and debunks in Spanish.

Figure \ref{fig:heatmap} displays the heatmap illustrating language dynamics of tweets and its related debunks in MMTweets. Notably, multiple languages exhibit near-zero associated debunks in languages besides English (e.g., Hindi), suggesting a potential gap in fact-checking coverage for specific languages. This emphasises the need to address disparities in debunk distribution and highlights opportunities for automated cross-language fact-checking methods like X-DNR.

\subsubsection{Temporal Diversity}
To assess dataset diversity, we also analyse the temporal characteristics of tweets in Figure \ref{fig:MMDate}, presenting a month-by-month breakdown of tweet counts for each language in MMTweets. We observe that Hindi and English tweets exhibit a relatively even distribution from Jan 2020 to Mar 2021. Conversely, Portuguese and Spanish tweets show a more concentrated presence, primarily emerging in late 2020 and early 2021. It's important to note that the MMTweets dataset encompasses at least one tweet for each month from Jan 2020 to Mar 2021 (spanning 15 months). Overall, we find the tweets to be temporally diverse across languages. 

% Distribution of Time Gaps between Tweet and Debunk
In examining cases where debunking precedes misinformation tweets (22.3\% of cases), Figure \ref{fig:gaps} illustrates publication date gaps. With a median gap of 76 days, the findings reveal misinformation can persist even after relevant debunks are available. For instance, one of the false tweets about ``Bill Gates launching implantable chips to track COVID-19,'' appeared in English on Twitter on 3 July 2020, while the earliest related debunk available was published on 13 May 2020
% \footnote{\url{https://factuel.afp.com/non-bill-gates-na-pas-propose-dimplanter-une-puce-electronique-la-population}}
\citep{afpbillgates} in the French language (49 days gap). This emphasises the need for effective methods, such as X-DNR, to detect the spread of already debunked narratives in multiple languages.

\subsubsection{Domain Diversity}
% Figure \ref{fig:MMWords} depicts the wordcloud for tweets for each language. These words represent events in the country where the language is spoken. As expected, the words related to coronavirus are apparent in all four languages. Some distinct words are specific to a given language. For instance, in Hindi tweets, words such as ``farmer'' and ``Delhi'' are related to the misinformation that spread during the farmers' protest in Delhi, India\footnote{\url{https://en.wikipedia.org/wiki/2020\%E2\%80\%932021_Indian_farmers\%27_protest}}. 
% In addition, words like ``China'' and ``Wuhan'' in English tweets, refer to the misinformation related to the origin of COVID-19. 
% The word ``vaccine'' is dominant in both Portuguese and Spanish misinformation tweets which is likely because the tweets for these languages are mainly from the end of 2020 (see Figure \ref{fig:MMDate}), when vaccine-related information was at its peak \cite{yousefinaghani2021analysis}. Overall, we find that the misinformation tweets in the MMTweets dataset are topically diverse. 

Table \ref{tab:lda} presents the results of topic modelling using Latent Dirichlet Allocation (LDA) \citep{blei2003latent}, showcasing the top three topics for each tweet language in MMTweets. As expected, the topics related to coronavirus are apparent in all four languages. However, some topics are specific to events in the country where the language is spoken. 
In Hindi, the first topic appears to focus on a combination of religious and political elements.
For instance, words such as ``farmer'' and ``Delhi'' are related to the misinformation that spread during the farmers' protest in Delhi, India
% \footnote{\url{https://en.wikipedia.org/wiki/2020\%E2\%80\%932021_Indian_farmers\%27_protest}}
\citep{wikifarmer}. Similarly, in Portuguese, the dominant topics revolve around President Bolsonaro and vaccines. The topics related to ``vaccine'' are dominant in both Portuguese and Spanish tweets which is likely because the tweets for these languages are mainly from the end of 2020 (see Figure \ref{fig:MMDate}), when vaccine-related information was at its peak \cite{yousefinaghani2021analysis}. English topics cover diverse aspects, including misinformation related to the origin of COVID-19 and its impact on people and hospitals. Overall, the table provides insights into the diverse and multifaceted nature of claims related to COVID-19 in MMTweets.

\begin{table}[]
\centering
\small
\caption{Topics captured by Latent Dirichlet Allocation.}
\resizebox{\columnwidth}{!}{%
\begin{tabular}{ll}
\toprule
\textbf{Language} & \textbf{Latent Dirichlet Allocation Topics}                                            \\ \midrule
Hindi      & Topic 1: hindu, delhi, corona, farmer, government    \\
Hindi       & Topic 2: going, people, world, temple, muslim           \\
Hindi       & Topic 3: massive, please, wipe, foreign, affair      \\
Portuguese       & Topic 1: people, bolsonaro, vaccine, world, covid    \\
Portuguese       & Topic 2: vaccine, work, mask, vote, without          \\
Portuguese       & Topic 3: vaccine, world, minister, took, abortion    \\
English       & Topic 1: deployed, mask, time, corona, epidemic      \\
English       & Topic 2: coronavirus, wuhan, like, china, year        \\
English       & Topic 3: people, work, coronavirus, hospital, covid \\
Spanish       & Topic 1 : people, without, vaccine, go, mask         \\
Spanish       & Topic 2: day, say, first, government, usa            \\
Spanish       & Topic 3: vaccine, died, nurse, covid, netherlands   \\ \bottomrule
\end{tabular}%
}
\label{tab:lda}
\end{table}

% The table presents the results of topic modeling using Latent Dirichlet Allocation (LDA) on text data in different languages, highlighting the top three topics for each language along with their respective counts. In Hindi (HI), the first topic (123 counts) appears to focus on a combination of religious and political elements, with terms such as 'hindu,' 'delhi,' 'corona,' 'farmer,' and 'government.' The second topic (114 counts) centers around general activities and observations with words like 'going,' 'people,' 'world,' 'temple,' and 'see.' The third topic (91 counts) seems to touch upon a more global or diplomatic discourse, featuring terms like 'massive,' 'please,' 'wipe,' 'foreign,' and 'affair.' Similarly, in Portuguese (PT), the dominant topics revolve around President Bolsonaro, vaccines, and global issues related to COVID-19. English (EN) topics cover diverse aspects, including military deployment, coronavirus origins, and the impact on people and hospitals. In Spanish (ES), topics involve discussions on vaccines, government statements, and notable events like the unfortunate death of a nurse in the Netherlands. This table provides a comprehensive snapshot of the prevalent themes in each language, offering insights into the multifaceted nature of public discourse surrounding the COVID-19 pandemic.

\section{Cross-lingual Debunked Narrative Retrieval (X-DNR)}
\label{debunkSearchMethods}

% X-DNR is formulated as an information retrieval (IR) task aimed at finding the most suitable debunked narrative across multiple languages based on a given query claim. In this paper, we exclusively focus on textual misinformation cases (totalling 1,176, as shown in Table \ref{tab:MMFlies}). Similar to prior work \citep{shaar-etal-2020-known, nakov2022overview}, we utilise tweets as queries to retrieve from a corpus of debunks. The ultimate goal of X-DNR is to provide the most accurate fact-checking information to users in response to potential misinformation claims in any language.

In this section, we formally define the X-DNR task. Given a tweet claim as a query $t$, the X-DNR system employs a retrieval model to obtain a candidate set of debunked narratives from a larger corpus of debunks $D = \{d_i\}_{i=1}^D$ in multiple languages. The final trained model can be expressed as $\text{X-DNR}(t, D)$, whose ultimate goal is to provide the most accurate fact-checking information to users in response to potential misinformation claims in any language.

In this paper, we exclusively focus on textual misinformation cases (totalling 1,176, as shown in Table \ref{tab:MMFlies}). For the retrieval corpus, we utilise a collection of $30,452$ previously gathered debunks in multiple languages (refer to Section \ref{MM Data Colletion}). Each debunk comprises a concatenated debunked claim and article title field (Section \ref{MM Data Colletion}).

\subsection{Cross-lingual Retreival Models}
\label{CLRmodels}

We test the following cross-lingual retrieval models on MMTweets.

\paragraph{\textbf{Okapi BM25.}} 
\label{elasticsearch}
We utilise the ElasticSearch \citep{gormley2015elasticsearch} implementation of BM25 \citep{gormley2015elasticsearch} with default parameters ($k = 1.2$ and $b = 0.75$). Since BM25 is designed for monolingual retrieval, we employ machine translation using the Fairseq's \texttt{m2m100\_418M} model \citep{fan2021beyond} to make it applicable to cross-lingual query and document pairs. All non-English tweets and debunks are translated into English, and the complete corpus of debunks is indexed in ElasticSearch \citep{gormley2015elasticsearch}. We then use the English-translated tweets as queries over the debunks.

% We use the ElasticSearch \citep{gormley2015elasticsearch}
% implementation of BM25~\citep{jones2000probabilistic}, with default parameters in ElasticSearch ($k = 1.2$ and $b = 0.75$). BM25 is a lexical-based retrieval method and only works for monolingual retrieval. Therefore we employ machine translation as a way of applying BM25 to cross-lingual query and document pairs. To this end, we translated all non-English tweets and debunks to English using the Fairseq's \texttt{m2m100\_418M} model \citep{fan2021beyond}. Then, we index the complete corpus of debunks in elasticsearch~\citep{gormley2015elasticsearch} and use the English-translated tweets as queries over the debunks.

\paragraph{\textbf{xDPR}}
\label{sec:xDPR}
% Dense Passage Retrieval (DPR) \citep{karpukhin2020dense} stands as one of the earliest dense retrieval models, employing BERT-based question and document encoders to evaluate document scores based on their similarity. However, DPR was initially designed to be initialised with English BERT. Therefore, we utilise a multilingual variant, namely xDPR
% % \footnote{\url{https://huggingface.co/eugene-yang/dpr-xlm-align-engtrained}}
% \citep{yang2022c3, hgxdpr}, which is an XLM-RoBERTa \citep{conneau2019unsupervised} model fine-tuned on the MSMARCO dataset \citep{NguyenRSGTMD16}. We further fine-tune the model on MMTweets using the same methodology \citep{yang2022c3}.

Dense Passage Retrieval (DPR) \citep{karpukhin2020dense}, an early dense retrieval model, uses BERT-based encoders for queries and documents to assess relevance based on their similarity. To expand its support beyond English, we use a multilingual variant, xDPR \citep{yang2022c3, hgxdpr}, which is an XLM-RoBERTa \citep{conneau2019unsupervised} model fine-tuned on the MSMARCO dataset \citep{NguyenRSGTMD16}. We further fine-tune xDPR on our MMTweets \citep{yang2022c3}.

\paragraph{\textbf{mContriever}}
\label{sec:mContriever}
% \citet{izacard2022unsupervised} introduced mContriever, which utilises contrastive loss to pretrain mBERT \citep{devlin-etal-2019-bert} on Wikipedia and CCNet \citep{wenzek2020ccnet} in an unsupervised manner. This method demonstrates improved performance in IR tasks. We utilised the authors’ provided multilingual checkpoint
% % \footnote{\url{https://huggingface.co/facebook/mcontriever-msmarco}}
% \citep{hgmcontriever}, which had already been fine-tuned on the MSMARCO dataset \citep{NguyenRSGTMD16} after the additional pretraining phase. We further fine-tune this model on MMTweets, employing the same methodology as described in \citet{izacard2022unsupervised}.

\citet{izacard2022unsupervised} introduced mContriever, which employs contrastive loss for unsupervised pretraining of mBERT \citep{devlin-etal-2019-bert}, showing enhanced performance on various IR tasks. We use the provided multilingual checkpoint \citep{hgmcontriever}, already fine-tuned on MSMARCO \citep{NguyenRSGTMD16}. We further fine-tune this model on MMTweets, employing the same methodology as described in \citet{izacard2022unsupervised}.

\paragraph{\textbf{Bi-Encoder (BE)}}
\label{sec:vmptmodel}
We fine-tune different Multilingual Pretrained Transformer (MPT) models as bi-encoders \citep{reimers2019sentence, karpukhin2020dense} on pairs of query tweets and their corresponding debunks. 
The objective function employed is the mean squared error, measuring the disparity between the true label and the model-calculated relevance score for the tweet-debunk pair. 
% This updates the model parameters such that the embedding of a misinformation tweet lies in proximity to its relevant debunks as compared to the other irrelevant debunks in the vector space. 
This adjusts model parameters, aligning the embedding of a query tweet closer to its relevant debunks in the vector space.
The loss equation is as follows,

\begin{equation}
% \small
\label{debunkscoreeq}
\mathcal{L}(\theta) = \frac{1}{\mathcal{N}} \sum_{i=1}^{\mathcal{N}} \left(\mathcal{Y}_i - \left(\frac{f_{\theta}(t_{i}) \cdot f_{\theta}(d_{i})}{\|f_{\theta}(t_{i})\|_{2}\|f_{\theta}(d_{i})\|_{2}}\right)\right)^{2}
\
\end{equation}

\noindent{where $f_{\theta}$ is the shared MPT encoder for tweet $t_{i}$ and debunk $d_{i}$,  \(\mathcal{Y}_i\) represents the true label of the \(i\)-th sample. The relevance score between tweet and debunk is computed using cosine similarity. 
% The embeddings are obtained via mean-pooling (i.e., by averaging embeddings of the constituent subwords of the input text). 
We employ cosine similarity with the mean-pooling technique due to its proven effectiveness in prior research \citep{reimers2019sentence}.  

We fine-tune bi-encoder using five different MPT models, namely multilingual BERT (mBERT) \citep{devlin-etal-2019-bert}, XLM-RoBERTa (XLMR) \citep{conneau2019unsupervised} and Language-Agnostic BERT Sentence Embedding (LaBSE) \citep{feng2020language}. Additionally, we also fine-tune two Sentence-Transformer model variants i.e. Universal Sentence Encoder (USE) \citep{hguse, yang2019multilingual}
% \footnote{USE variant which supports around 50 languages. Link: \url{https://huggingface.co/sentence-transformers/distiluse-base-multilingual-cased-v2}}
and Masked and Permuted Pretraining for Language Understanding (MPNet)
% \footnote{Link: \url{https://huggingface.co/sentence-transformers/paraphrase-multilingual-mpnet-base-v2}}
\citep{hgmpnet, song2020mpnet}. These bi-encoder models are denoted by the prefix ``BE-'' 
in subsequent experiments (Section \ref{CLRADNresult1}). 
% The hyperparameter details are mentioned in Appendix \ref{appendix:hyper}.

\paragraph{\textbf{Multistage Retrieval}}
\label{MultistageRetrieval}
Drawing inspiration from the success of multistage retrieval methods in IR tasks \citep{nogueira2019passage, thakur2021beir, singh2021multistage}, we apply these techniques to the X-DNR task. Within this context, we introduce two methods that adapt earlier approaches, specifically tailored for the X-DNR task. These methods are as follows:

\begin{itemize}[noitemsep, leftmargin=*]
\setlength\itemsep{0.3em}
    \item \textbf{Bi-Encoder+Cross-Encoder (\textit{BE+CE}):} In the first retrieval stage, we fine-tune an MPT model as a bi-encoder instead of the standard BM25-based lexical retrieval approach adopted in prior work \citep{shaar-etal-2020-known, kazemi2021claim}. This choice is motivated by the MPT model's suitability for the cross-lingual nature of the task, eliminating the need for translation. 
    In the second stage, we fine-tune an MPT model as a cross-encoder \citep{nogueira2019passage} to re-rank the top-$K$ retrieved debunks from the first stage. Here, the model employs self-attention mechanisms on the given tweet and debunk pair to get the final relevance score. The input to the model follows the structure: 
    $ 
        [CLS]~~[T_{1}]...[T_{n}]~~[SEP]~~[DC_{1}]...[DC_{i}][DT_{1}]...[DT_{j}],~~
    $
    where $T_{n}$ are the tweet subword tokens and $DC_{i}$ and $DT_{j}$ are the debunked claim and title subword tokens, respectively. $[CLS]$ and $[SEP]$ are the default tokens to indicate “start of input” and “separator”, respectively, in the Next Sentence Prediction task \citep{devlin-etal-2019-bert}. 
    % Please refer to Appendix \ref{appendix:hyper} for hyperparameter details.

    \item \textbf{Bi-Encoder+ChatGPT (\textit{BE+GPT3.5}):} Large language models like ChatGPT (\textit{gpt-3.5-turbo}) have consistently showcased impressive capabilities across a broad spectrum of natural language processing tasks
    % \footnote{\url{https://platform.openai.com/docs/models}}. 
    \citep{openai}.
    % However, their utilisation in IR tasks remains an active area of exploration, aiming to enhance their ability to retrieve relevant documents from large corpus in response to a specific input query.
    Therefore, to evaluate ChatGPT's performance, we implement a Listwise Re-ranker with a Large Language Model (LRL) \citep{ma2023zero} to re-rank documents retrieved by the first stage ranker. The main distinctions in our approach compared to prior work \citep{ma2023zero} are: 1) we employ multilingual bi-encoders described in Section \ref{sec:vmptmodel} as the first-stage ranker 2) each re-ranked document consists of concatenated debunk claim and title fields. Besides this, all parameters are kept same as used by \citet{ma2023zero}.
\end{itemize}

\subsection{Experimental Details}
\label{appendix:hyper}

\subsubsection{Train and test sets}
We divide 1,176 textual misinformation tweet queries into train and test sets. The test set consists of $400$ tweet queries ($100$ queries per language), comprising the same triple-annotated tweets used for calculating IAA (Section \ref{statistics}). The remaining 776 tweet queries are used as training data, with a 10\% subset used as a validation set. Please note that during test time, we do not know if a tweet has been debunked, because tweets linked with debunks in the test set do not occur in the train set. This ensures a realistic test scenario by preventing tweets linked to the same debunk from appearing in both the train and test sets.

% The final training dataset comprises 10,120 tweet and debunk pairs, with 2,360 positive (1,420 exact matches and 940 partial matches) and 7,760 negative tweet and debunk pairs. 
Now, since each query tweet in the training set is linked to multiple debunks (Section \ref{claimmatchannotate}), therefore, the final training set comprises 2,360 positive (1,420 exact matches and 940 partial matches) tweet and debunk pairs. For negative pairs, ten debunks are randomly sampled for each tweet, resulting in total 7,760 negative tweet and debunk pairs. 
% For a comprehensive analysis of various methods used for getting negative tweet and debunk pairs, please refer to Section \ref{CLRADNresult4}. 
We also experimented with hard negative mining and higher counts of negatives, but did not observe any significant improvements.
In total, the training set consists of 10,120 fine-grained tweet and debunk pairs for training different retrieval models.

\subsubsection{Evaluation Metrics} 
We employ two widely used ranking metrics \citep{nakov2021clef, nakov2022overview} for evaluation: Mean Reciprocal Rank (\textbf{MRR}) and Normalised Discounted Cumulative Gain (\textbf{nDCG@1} \& \textbf{nDCG@5}).
MRR measures the effectiveness of the system by computing the score based on the highest-ranked relevant debunk for each misinformation tweet. MRR is defined as 
$
    \mathrm{MRR}=\frac{1}{|\mathcal{T}|} \sum_{i=1}^{|\mathcal{T}|} \frac{1}{rank}_{i},
% \end{equation}
$
where $|\mathcal{T}|$ is the number of tweets used as query and $rank_{i}$ is the rank of the top relevant debunk for the $ith$ tweet.
On the other hand, the nDCG@K normalises DCG@K by dividing it by ideal DCG@K, where DCG@K discounts the graded relevance value of retrieved debunks based on their ranks. DCG@K is defined as follows,
\begin{equation}
DCG@K=\frac{1}{|\mathcal{T}|} \sum_{i=1}^{|\mathcal{T}|} \sum_{k=1}^{|\mathcal{K}|} \frac{2^{rel_{i, k}} - 1}{\log_2(rank_{i, k}+1)},
\end{equation}
% \begin{equation}
%     \label{eq:ndcg}
%     nDCG@K=\frac{DCG@K}{iDCG@K},
% \end{equation}
where $rel_{i, k}$ is the graded relevance of the debunk at $rank_{i, k}$ for the $i$th query tweet. Higher MRR and nDCG scores indicate better performance.

\subsubsection{Hyperparameters}
The bi-encoder is trained for four epochs with a batch size of $32$, a learning rate of $4e-5$ and maximal input sequence length of $256$. The cross-encoder, trained for two epochs, uses a batch size of $16$, $4e-5$ learning rate, with truncation of subword tokens beyond $512$. Both models employ linear warmup, AdamW optimiser, and manual hyperparameter tuning on a validation set. Hyperparameter bounds are set as: 1) $1$ to $5$ epoch 2) $1e-5$ to $5e-5$ learning rate 3) $8$ to $64$ batch size on NVIDIA RTX 3090. 
% All experiments are conducted on NVIDIA RTX 3090.

% \subsection{Evaluation Measures} 

% We use Mean Reciprocal Rank (MRR) and Mean Average Precision (MAP) as ranking evaluation metrics in this study.
% % These metrics have been widely adopted in previous studies \citep{shaar-etal-2020-known, nakov2021clef}.
% MRR measures the effectiveness of the system by computing the score based on the highest-ranked relevant debunk for each misinformation tweet. MRR is defined as 
% $
%     \mathrm{MRR}=\frac{1}{|T|} \sum_{i=1}^{|T|} \frac{1}{\operatorname{rank}_{i}},
% % \end{equation}
% $
% where $|T|$ is the number of tweets used as query and $rank_{i}$ is the rank of the relevant debunk for the $ith$ tweet. The higher the MRR score the better.
% % \textcolor{blue}{
% MAP, on the other hand, measures the precision of the system in returning relevant results for a given query. 
% % MAP is computed by taking the mean of average precision scores for each query. 
% We use two variations of MAP: MAP@1 and MAP@5, which evaluate the top one and top five retrieved documents, respectively. A higher MAP@k score indicates better performance.
% % }

% We employ two widely used ranking evaluation metrics \citep{nakov2021clef, nakov2022overview} for evaluation
% %are selected to evaluate the performance of our systems and baselines
% : Mean Reciprocal Rank (\textbf{MRR}) and Normalized Discounted Cumulative Gain (\textbf{nDCG@1} \& \textbf{nDCG@5}).

\begin{table*}[]
\centering
\small
\caption{Results for different cross-lingual retrieval models on the test set of MMTweets. The best scores are in bold. }
\begin{tabular}{@{}llcccccccccc@{}}
\toprule
\textbf{Language}    & \textbf{Metric} & \textbf{BM25} & \textbf{xDPR} & \textbf{mCont} & \textbf{BE-mBERT} & \textbf{BE-XLMR} & \textbf{BE-USE} & \textbf{BE-LaBSE} & \textbf{BE-MPNet} & \textbf{BE+CE} & \textbf{BE+GPT3.5} \\ \midrule
\textbf{MMTweets-HI} & \textbf{nDCG@1} & 0.263         & 0.435                 & 0.240                        & 0.135             & 0.160             & 0.210           & 0.525             & 0.320             & \textbf{0.610} & 0.575              \\
                     & \textbf{nDCG@5} & 0.267         & 0.421                 & 0.304                        & 0.149             & 0.188             & 0.246           & 0.514             & 0.366             & \textbf{0.569} & 0.527              \\
                     & \textbf{MRR}    & 0.320         & 0.503                 & 0.352                        & 0.199             & 0.250             & 0.310           & 0.623             & 0.439             & \textbf{0.674} & 0.637              \\
\textbf{MMTweets-PT} & \textbf{nDCG@1} & 0.625         & 0.695                 & 0.770                        & 0.540             & 0.685             & 0.730           & 0.755             & 0.755             & \textbf{0.845} & 0.840              \\
                     & \textbf{nDCG@5} & 0.598         & 0.690                 & 0.761                        & 0.514             & 0.595             & 0.672           & 0.726             & 0.720             & \textbf{0.765} & 0.757              \\
                     & \textbf{MRR}    & 0.723         & 0.781                 & 0.849                        & 0.627             & 0.737             & 0.782           & 0.822             & 0.821             & \textbf{0.887} & 0.880              \\
\textbf{MMTweets-EN} & \textbf{nDCG@1} & 0.591         & 0.635                 & 0.705                        & 0.515             & 0.465             & 0.675           & 0.680             & 0.710             & \textbf{0.720} & 0.715              \\
                     & \textbf{nDCG@5} & 0.572         & 0.625                 & 0.670                        & 0.475             & 0.472             & 0.638           & 0.650             & \textbf{0.696}    & 0.682          & 0.662              \\
                     & \textbf{MRR}    & 0.706         & 0.759                 & 0.801                        & 0.603             & 0.590             & 0.760           & 0.780             & \textbf{0.814}    & \textbf{0.814} & 0.807              \\
\textbf{MMTweets-ES} & \textbf{nDCG@1} & 0.560         & 0.620                 & 0.610                        & 0.405             & 0.435             & 0.500           & 0.585             & 0.615             & \textbf{0.735} & 0.660              \\
                     & \textbf{nDCG@5} & 0.525         & 0.621                 & 0.646                        & 0.394             & 0.428             & 0.497           & 0.582             & 0.582             & \textbf{0.662} & 0.632              \\
                     & \textbf{MRR}    & 0.648         & 0.707                 & 0.730                        & 0.491             & 0.536             & 0.591           & 0.670             & 0.678             & \textbf{0.804} & 0.741              \\ \midrule
\textbf{Average}     & \textbf{nDCG@1} & 0.510         & 0.596                 & 0.581                        & 0.399             & 0.436             & 0.529           & 0.636             & 0.600             & \textbf{0.728} & 0.698              \\
                     & \textbf{nDCG@5} & 0.490         & 0.589                 & 0.595                        & 0.383             & 0.421             & 0.513           & 0.618             & 0.591             & \textbf{0.669} & 0.644              \\
                     & \textbf{MRR}    & 0.599         & 0.687                 & 0.683                        & 0.480             & 0.528             & 0.611           & 0.724             & 0.688             & \textbf{0.795} & 0.766              \\ \bottomrule
\end{tabular}

\label{tab:DebRes1}
\end{table*}

\section{Results and Discussion}
\label{resandiscuss}

In this section, we present the results of retrieval experiments that aim to address the following five research questions:

\begin{itemize}[noitemsep,leftmargin=9.5mm]
    \item[\textbf{RQ1}]{To what extent do the current SOTA cross-lingual retrieval models perform in addressing the specific challenges posed by the MMTweets dataset? (Section \ref{CLRADNresult1})}
    \item[\textbf{RQ2}]{How challenging is it for models to transfer and generalise across languages within MMTweets? (Section \ref{CLRADNresult2})}
    \item[\textbf{RQ3}]{Can models trained on existing datasets transfer knowledge and generalise on the MMTweets test set? (Section \ref{CLRADNresult3})}
    % \item[\textbf{RQ4}]{How does the type and count of negative pairs impact the model's performance on MMTweets? (Section \ref{CLRADNresult4})}
    \item[\textbf{RQ4}]{What insights can be gained into the retrieval latency of various cross-lingual retrieval models? (Section \ref{CLRADNresult5})}
\end{itemize}

\subsection{Model Performance}
\label{CLRADNresult1}

Table \ref{tab:DebRes1} shows Mean Reciprocal Rank (MRR) and Normalised Discounted Cumulative Gain (nDCG@1 \& nDCG@5) on the test set of MMTweets (HI, PT, EN \& ES). 
The results suggest that BE-mBERT and BE-XLMR consistently show lower scores, with occasional lower performance when compared to BM25. BM25's strength lies in lexical overlap with machine-translated text, giving it an advantage over other models.
However, other retrieval models outperform BM25 on several metrics. Notably, BE-LaBSE performs better than BE-MPNet, BE-USE, BE-mBERT, and BE-XLMR, even outperforming state-of-the-art models like xDPR and mContriever in average metric scores. This is attributed to LaBSE's sentence-level objective, combined with pretraining techniques involving translation and masked language modelling, as discussed in \citet{feng2020language}.

The last two columns of Table \ref{tab:DebRes1} report the scores of multistage retrieval methods (\textit{BE+CE} \& \textit{BE+GPT3.5}). In multistage retrieval, we employ LaBSE for the first stage due to its superior performance over other models (see Table \ref{tab:DebRes1}). Similarly, the second stage in \textit{BE+CE} also utilises LaBSE, with the number of re-ranked documents set to 20. Although we experimented with various MPT models and different counts of re-ranked documents in the second stage, no significant improvements were observed. 
The results show that \textit{BE+CE} consistently emerges as the top performer across all datasets and metrics, achieving an average nDCG@1 score of 0.728, an average nDCG@5 score of 0.669, and an average MRR score of 0.795 (Table \ref{tab:DebRes1} -- second last column). On the other hand, while \textit{BE+GPT3.5} outperforms other models in average metric scores, its retrieval latency is the highest (see section \ref{CLRADNresult5}). 
% In general, we find that multistage retrieval is effective for X-DNR as compared to other models.
% Furthermore, it is interesting to note that the effectiveness of \textit{BE+CE} is consistent across metrics. 
Although other models like BE-LaBSE, BE-MPNet, xDPR, and mContriever showcase competitive performance, none consistently match the performance demonstrated by multistage retrieval methods. Additionally, despite being trained on the extensive MSMARCO training dataset (Section \ref{CLRmodels}), models such as xDPR and mContriever do not notably enhance performance, suggesting distinctive challenges presented by MMTweets.

% Overall, \textit{BE+CE} improves the results, with an average increase of 43\% in nDCG@1, 37\% in nDCG@5, and 33\% in MRR compared to BM25. In comparison to BE-LaBSE, \textit{BE+CE} has an average increase of 14\% for nDCG@1, 8\% for nDCG@5, and 10\% for MRR metric. 

For \textit{BE+CE}, the extent of improvement varies across languages. For example, in the case of Portuguese, \textit{BE+CE} outperforms BM25 with increases of 132\% for nDCG@1, 112\% for nDCG@5, and 110\% for MRR. Conversely, the improvement is relatively low for English, with increases of only 22\%, 19\%, and 15\% for nDCG@1, nDCG@5, and MRR scores, respectively. We hypothesise that this disparity in performance across different languages may be attributed to noisy translations in the case of BM25, while \textit{BE+CE} doesn't rely on translation. Additionally, the scores on different languages are reflective of how topics found in each language impact a model's performance. For example, the Hindi tweets have the lowest performance across all models and evaluation metrics, which suggests that the topics found in these languages (Section \ref{statistics}) are quite challenging for the model. Another reason for poor Hindi performance could be the change in the language script to Devanagari. This suggests that dealing with various languages in MMTweets poses a challenge, and there is still potential for improvement in retrieval models.

Furthermore, we also observe the challenge of distinguishing closely related debunks by the model. This occurs when the retrieved debunk is not entirely relevant, but still shares some degree of relevance with the query claim. For instance, consider the query claim about the sighting of crocodiles in the flooded streets of Hyderabad; the top-retrieved debunks are closely related, involving sightings of crocodiles in Mumbai, Bengaluru, Florida, etc. This highlights the need for continued refinement in retrieval models to enhance the relevance of top-ranked debunks for the X-DNR task.

In summary, these evaluations highlight performance differences among models, emphasising the consistent superiority of multistage retrieval methods across various languages and metrics. While BM25 is faster (see Section \ref{CLRADNresult5}), the necessity of machine translation for BM25 incurs additional costs and time overheads.

% In summary, these evaluations highlight performance disparities among the models, emphasising the consistent superiority of the multistage retrieval methods (\textit{BE+CE} \& \textit{BE+GPT3.5}) across different languages and metrics. Although BM25 is faster than the multistage retrieval methods (see Section \ref{CLRADNresult5}), the need to machine-translate data for BM25 introduces additional costs and time overheads.
% \footnote{For instance, the machine translation model we used in Section \ref{elasticsearch} operates at 22 sentences per second on a V100 GPU.}. 
% Conversely, the neural retrieval methods can be optimised by caching the embeddings. 

% While BE+ChatGPT3.5 outperforms other models in average metric scores, the performance of BE+ChatGPT3.5 varies depending on the dataset and evaluation metric. For example, on the MMTweets-HI dataset, BE+ChatGPT3.5 achieves second highest MRR score of 0.637, while BM25 achieves a score of only 0.61. On the other hand, on the MMTweets-ES dataset, BE+ChatGPT3.5 achieves the MAP@5 score of 0.72, while BM25 achieves the highest score of 0.75.
% % Additionally, variations in the structure of titles and claims across different fact-checking websites within our corpus could also contribute to these differences. 

\begin{figure}[]
    \centering
\scalebox{0.7}{    \includegraphics[width=9.5cm,height=5cm]{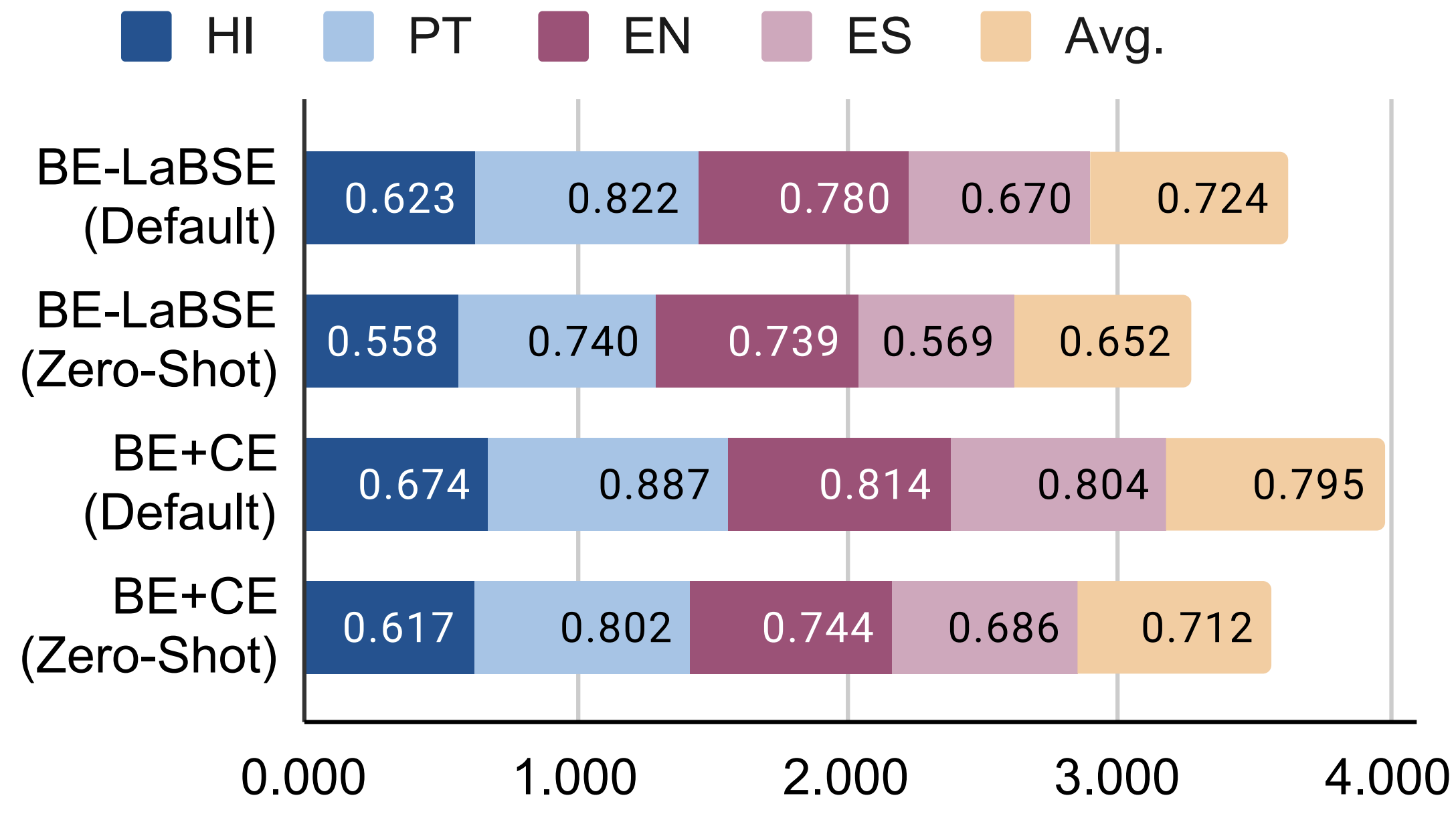}
}
    % \vspace{-5pt}
    \caption{Stacked bar plot for MRR scores for zero-shot cross-lingual transfer and the default results (from Table \ref{tab:DebRes1}).}
    % \caption{Cross-lingual retrieval of already debunked narratives (CLRADN): Hindi tweet and English relevant debunk.}

    % \vspace{-10pt}
    \label{fig:cross-ling}
\end{figure}

\subsection{Cross-lingual Transfer}
\label{CLRADNresult2}

To test the zero-shot transfer capabilities, the model is trained on languages other than the one it is tested on. For instance, to test zero-shot transfer for Hindi, the models are trained on only those tweet and debunk pairs that are not in Hindi. Hence, in total four models are trained for four different languages in the MMTweets.

We evaluate the cross-lingual transfer capability of BE-LaBSE and \textit{BE+CE}, which yield the highest average scores (Table \ref{tab:DebRes1}). 
Figure \ref{fig:cross-ling} shows a stacked bar plot illustrating MRR scores for zero-shot cross-lingual transfer and the default results sourced from Table \ref{tab:DebRes1}. 

When comparing the zero-shot results with the default results, the default results consistently outperform zero-shot results for both models (BE-LaBSE and \textit{BE+CE}) across all languages, as expected due to training on the complete dataset. Nevertheless, zero-shot models surpass several baselines, including BM25 (from Table \ref{tab:DebRes1}) in this challenging setting. The results suggest that models have the potential to transfer knowledge between languages without the need for language-specific training. This also supports prior observations that MPT models, when fine-tuned on monolingual data, exhibit strong performance on a different language \citep{izacard2022unsupervised, conneau2018xnli}. Despite these promising outcomes, there is still room for improvement for zero-shot models to match the performance of default models.

\begin{table}[]
\centering
\caption{Domain overlap between the test set of MMTweets and the train set of other datasets.}
 \label{tab:domoverlap}
\resizebox{\columnwidth}{!}{%
\begin{tabular}{@{}lcccccc
>{\columncolor[HTML]{FFFFFF}}c @{}}
\toprule
\textbf{Train Set} & \textbf{MMTweets}            & \textbf{Snopes}              & \textbf{CLEF 20-EN}          & \textbf{CLEF 21-EN}          & \textbf{CLEF 21-AR}          & \textbf{CLEF 22-EN}          & \textbf{CLEF 22-AR}          \\ \midrule
\textbf{Overalp}   & \cellcolor[HTML]{57BB8A}0.29 & \cellcolor[HTML]{EEF8F3}0.15 & \cellcolor[HTML]{FFFFFF}0.14 & \cellcolor[HTML]{F2BEB9}0.12 & \cellcolor[HTML]{E7F5EE}0.16 & \cellcolor[HTML]{E67C73}0.11 & \cellcolor[HTML]{FDF6F6}0.13 \\ \bottomrule
\end{tabular}%
}
\end{table}

\begin{figure*}%
    \centering
    \subfloat[\centering BE-LaBSE]{{\includegraphics[width=8.5cm,height=2.6cm]{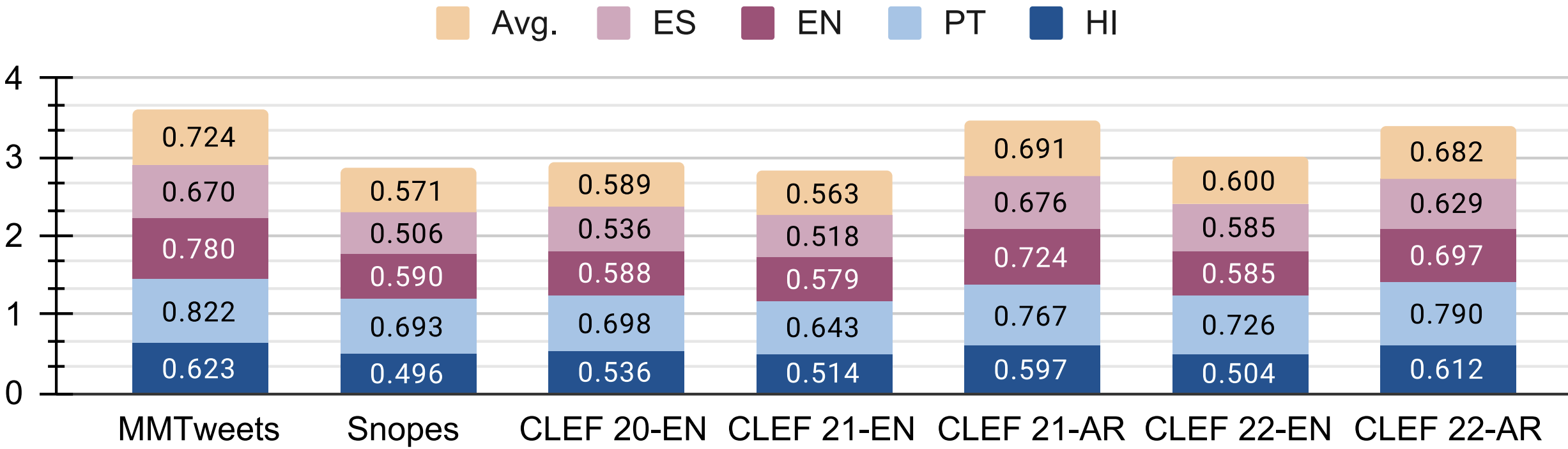} }}%
    \qquad
    \subfloat[\centering BE+CE]{{\includegraphics[width=8.5cm,height=2.6cm]{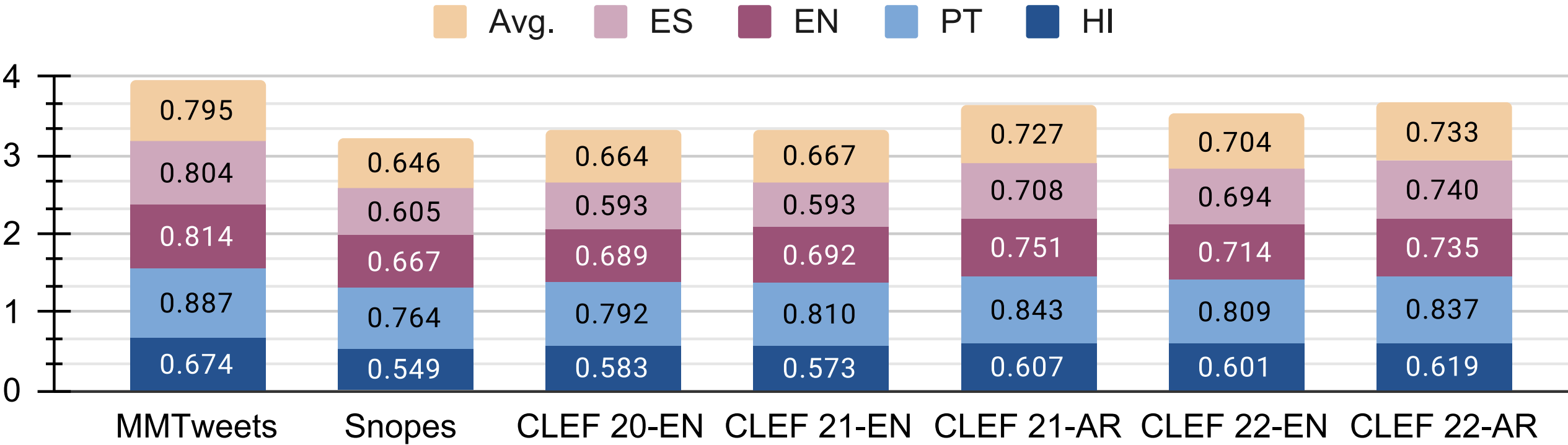} }}%
    \caption{Stacked bar plot for MRR scores for zero-shot cross-dataset transfer using BE-LaBSE (a) and BE+CE (b).}%
    \label{fig:crossdata}%
\end{figure*}

\subsection{Cross-dataset Transfer}
\label{CLRADNresult3}

To test the zero-shot cross-dataset transfer capabilities of the models, we train them on the training set of previously published datasets and subsequently evaluate their performance on the test set of MMTweets. This ensures real-life testing to assess the generalisability of the models. The previously published datasets include Snopes \cite{shaar-etal-2020-known} and CLEF CheckThat! Lab task datasets which include CLEF 22-EN and CLEF 22-AR \cite{nakov2022overview}, CLEF 21-EN and CLEF 21-AR, \cite{nakov2021clef} and CLEF 20-EN \cite{shaar2020overview}. Please note that CLEF 22-AR and CLEF 21-AR are Arabic datasets while other datasets are in English.

First, we assess the domain overlap to see how challenging it is for models trained on existing datasets to transfer knowledge to the MMTweets test set. For this, we use weighted Jaccard similarity \cite{ioffe2010improved} to compute the domain overlap between the test set of MMTweets and the train set of other datasets used for cross-dataset analysis (Table \ref{tab:domoverlap}). We also report the overlap between the train and test set of MMTweets for reference. We find low domain overlap (ranging from 11-16\%) with other datasets' train sets compared to MMTweets' train set (which has a 29\% overlap) indicating distinct or less common instances between MMTweets and other datasets. We also conducted this analysis for each language but didn't find much variation in the results. Overall, MMTweets stands out as a unique dataset, showing low domain overlap with existing datasets.

Figure \ref{fig:crossdata} shows MRR scores for zero-shot cross-dataset transfer using BE-LaBSE (a) and \textit{BE+CE} (b), alongside default MMTweets trained results (from Table \ref{tab:DebRes1}). Notably, models trained on CLEF 21-AR and CLEF 22-AR, despite being in Arabic, achieve the highest scores across all languages after the default MMTweets trained models.
Additionally, models fine-tuned on CLEF 22-EN and 20-EN closely compete with other retrieval models (Table \ref{tab:DebRes1}). Notably, while all claims in other datasets are either in English or Arabic, the MMTweets test set encompasses multiple other languages, making it even more challenging to retrieve the best matching debunk. 

Overall, the findings suggest some knowledge transfer between datasets, which is especially valuable when obtaining a domain-specific dataset for training a dedicated model is challenging. However, despite these positive outcomes, there remains potential for models to match or surpass default MMTweets trained results.

\begin{figure}[]
    \centering
\scalebox{0.7}{    \includegraphics[width=8cm,height=4.5cm]{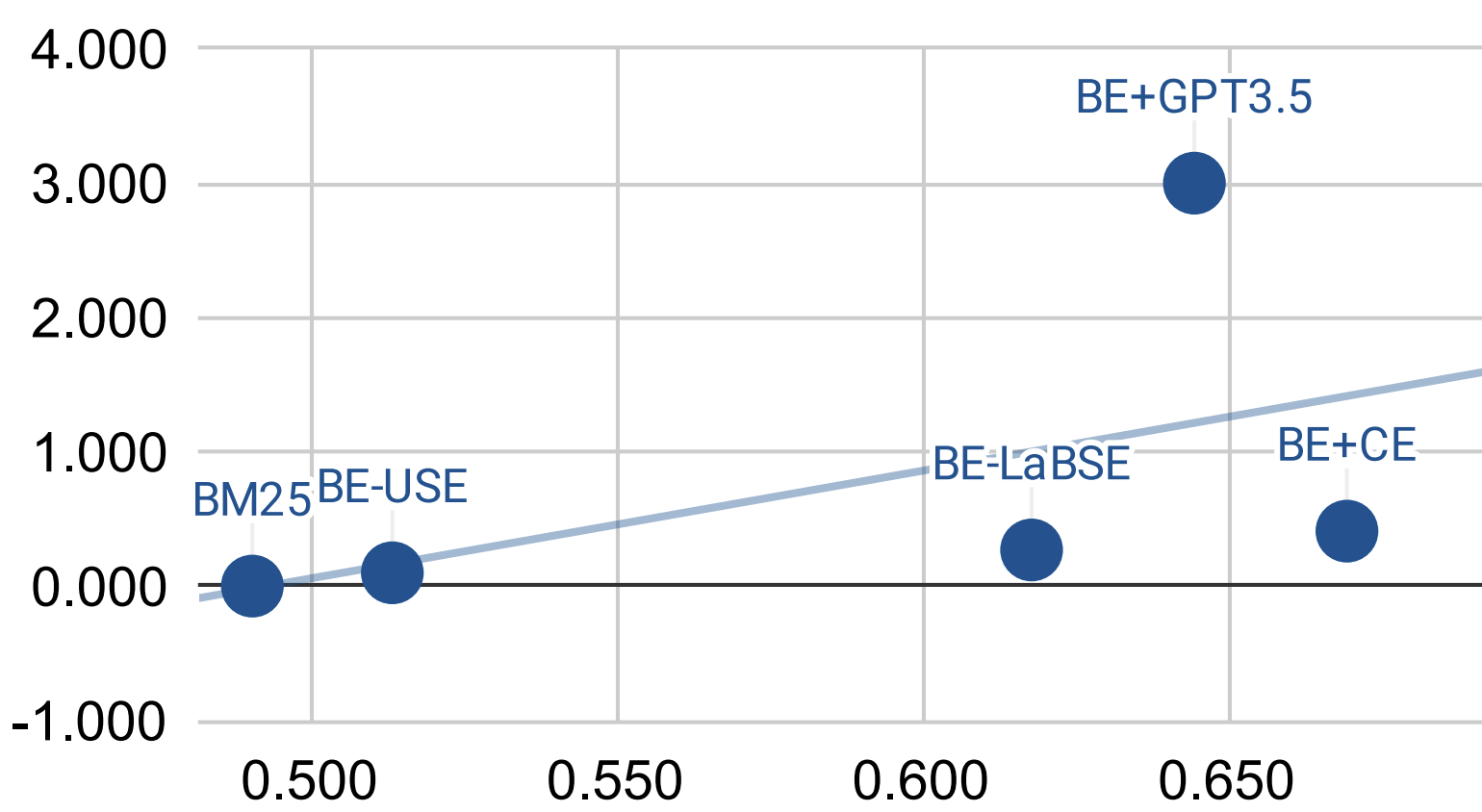}
}
    % \vspace{-5pt}
    \caption{Retrieval latency (in seconds) and MRR scores.}
    % \caption{Cross-lingual retrieval of already debunked narratives (CLRADN): Hindi tweet and English relevant debunk.}

    % \vspace{-10pt}
    \label{fig:time}
\end{figure}

\subsection{Retrieval Latency}
\label{CLRADNresult5}

Figure \ref{fig:time} shows scatter plot for the average MRR scores and retrieval latency for different models. Lower values in the retrieval latency indicate faster query processing by the IR model.
When comparing models, \textit{BE+CE} achieves the highest MRR score (0.669) but exhibits a latency of 0.41 seconds, indicating a comparatively longer retrieval time. BE-LaBSE follows closely with an MRR score of 0.618 and a moderate retrieval latency of 0.27 seconds, striking a balance between performance and retrieval speed. While \textit{BE+GPT3.5} displays a competitive MRR score (0.644), its retrieval latency increases to 3 seconds, impacting its practical application in real-time scenarios. BM25, although has the fastest retrieval latency at 0.001 seconds, it compromises ranking quality with the lowest MRR score of 0.490. 

Overall, BE-LaBSE provides a balanced option with reasonable performance and moderate retrieval latency, while \textit{BE+CE} excels in ranking quality, albeit with a slightly longer retrieval latency.

\section{Conclusion and Future Work}
\label{debunksearchconclusion}

This paper focuses on cross-lingual debunked narrative retrieval (X-DNR) for automated fact-checking. It introduces MMTweets, a novel benchmark dataset that stands out, featuring cross-lingual pairs, human annotations, fine-grained labels, and images, making it a comprehensive resource compared to other datasets. Furthermore, initial tests benchmarking SOTA cross-lingual retrieval models reveal that dealing with multiple languages in the MMTweets dataset poses a challenge, indicating a need for further improvement in retrieval models. Nevertheless, the introduction of tailored multistage retrieval methods demonstrates superior performance over other SOTA models, achieving an average nDCG@5 of 0.669. However, it's crucial to note the trade-offs between model performance and retrieval latency, with \textit{BE+CE} offering better ranking quality at the expense of longer retrieval times. Finally, the findings also suggest some knowledge transfer across languages and datasets, which is especially valuable in scenarios where language-specific models are not available or feasible to train. However, despite these positive outcomes, there is still room for models to match or even surpass the performance of default MMTweets trained models. To achieve this objective, the model needs an in-depth understanding of both language and context, along with the capability to differentiate among closely related debunked narratives. More sophisticated models could potentially introduce these capabilities in the future.

In future, we plan to extend the dataset to include claims from other social media platforms and domains to enhance its generalisability. Additionally, we aim to explore multimodal debunked narrative retrieval, leveraging information from various modalities.

% Finally, MMTweets is released as a benchmark dataset and we invite fellow researchers to evaluate their retrieval models on the presented dataset. 

\begin{acks}
This research is supported by a UKRI grant EP/W011212/1, an EU Horizon 2020 grant (agreement no.871042) (“SoBigData++: European Integrated Infrastructure for Social Mining and BigData Analytics” (\url{http://www.sobigdata.eu})) and a European Union grant INEA/CEF/ICT/A2020/2381686 (European Digital Media Observatory (EDMO) Ireland, https://edmohub.ie/).
\end{acks}
\bibliographystyle{ACM-Reference-Format}
\bibliography{sample-base}

\end{document}

% \title{Supplementary Material}
% \maketitle

\section{Appendix}

\begin{table*}[!htbp]
\centering
\small
\caption{English translations of the top six most frequent words between January 2020 and March 2021, grouped into quarters per language.}
\begin{tabular}{p{0.1\linewidth}p{0.15\linewidth}p{0.15\linewidth}p{0.15\linewidth}p{0.15\linewidth}p{0.15\linewidth}}
\toprule
\multicolumn{1}{c}{\multirow{2}{*}{\textbf{Language}}} & \multicolumn{5}{c}{\textbf{Month}} \\  \cline{2-6}
\multicolumn{1}{c}{} & \textbf{Jan-Mar 2020} & \textbf{April-Jun 2020}
 & \textbf{July-Sept 2020} & \textbf{Oct-Dec 2020}
 & \textbf{Jan-Mar 2021} \\ \midrule
\textbf{Hindi} & hindu, world, murder, police, delhi, people & death, corona, world, people, virus, religion & foreign, affairs, temple, hindu, lion, country & government, farmers, country, modi, supreme, hindu & city, india, farmers, people, father, movement \\ \midrule
\textbf{Portuguese} & bolsonaro, water, mercury, venus, saturn, pyramids & lions, streets, supermarket, carnival, bahia, government & vaccine, world, covid, minister, france, palestine & vaccine, people, trump, years, votes, covid & people, vaccine, world, masks, health, woman \\ \midrule
\textbf{English} & coronavirus, wuhan, china, patients, war, ronaldo & japan, coronavirus, epidemic, professor, italy, nobel & covid, mask, mother, kali, people, video, violence & passport, singapore, mask, pharaoh, crocodile, hyderabad & myanmar, pakistan, military, woman, khan, news, coup \\ \midrule
\textbf{Spanish} & tarragona, explosion, day, petrochemical, girl, election & government, people, coronavirus, world, order, health & years, vaccine, age, spanish, influenza, deaths & october, vaccine, netherlands, nurse, cnn, video & vaccine, covid, people, spain, minister, trump \\ \bottomrule
\end{tabular}
% \caption{English translations of top six most frequent words from January 2020 to March 2021. These are grouped into three months each for all languages in the MMTweets dataset.}
% \caption{English translations of the top six most frequent words between January 2020 and March 2021, grouped into quarters per language.}

\label{tab:MMWords2}
\end{table*}

\subsection{Word Distribution}
\label{app:worddist}

Table \ref{tab:MMWords2} shows the English translations of the top most frequent words per month. For a better understanding, these are grouped into three months each. Although our dataset spans from January 2020 to March 2021, we find that the tweets are topically diverse for each language. For instance, in Hindi tweets, misinformation related to the farmers' protest in Delhi\footnote{\url{https://en.wikipedia.org/wiki/2020\%E2\%80\%932021_Indian_farmers\%27_protest}} is highly concentrated after October 2020. In addition, misinformation related to the Ayodhya temple land dispute\footnote{\url{https://en.wikipedia.org/wiki/Ayodhya_dispute}} is rampant between July and September 2020. We also find that vaccine misinformation is dominant after July 2020 in both Portuguese and Spanish misinformation tweets. The English misinformation tweets mostly contain COVID-19 related misinformation until the end of 2020, however, it's different from January to March 2021 (Table \ref{tab:MMWords2}).

\subsection{Hyperparameters}
\label{appendix:hyper}

This section presents the training details. The first stage model is trained for four epochs with a batch size of $16$, a learning rate of $4e-5$ and maximal input sequence length of $512$. For the second stage, the cross-encoder model is trained with a batch size of $16$ and $4e-5$ learning rate for two epochs. The subword tokens beyond $512$ are truncated. For training the second stage model, we randomly sample ten negative debunks for each misinformation tweet. 
For all the models, we use linear warmup as the learning rate scheduler and AdamW as optimiser. 
The models are validated using the MMTweets training set and we manually tune the hyperparameters. The bounds for each hyperparameter are as follows: 1) 1 to 5 epoch 2) 1e-5 to 5e-5 learning rate 3) 8 to 64 batch size which is limited to model's GPU requirement. The training time for each epoch in first and second retrieval stage is 10 and 15 minutes respectively. 
% Since, we fine-tune the models, therefore, the initial training parameters for MPT models is same as the ones stated by authors in their original papers \citep{feng2020language, devlin-etal-2019-bert}.
All experiments are conducted on a machine with NVIDIA GeForce RTX 3090.

\begin{table}[]
% \small
\centering
\caption{Results of BM25 and off-the-shelf MPT models.}
\resizebox{\columnwidth}{!}{%
\begin{tabular}{@{}llccccc@{}}
\toprule
\textbf{Dataset}     & \textbf{Metrics} & \textbf{BM25} & \textbf{mBERT} & \textbf{XLM-RoBERTa} & \textbf{LaBSE} & \textbf{USE} \\ \midrule
\textbf{MMTweets-HI} & \textbf{MAP@1}   & \textbf{0.55}          & 0.01          & 0.00                 & 0.45           & 0.27         \\
\textbf{}            & \textbf{MAP@5}   & \textbf{0.61}          & 0.02          & 0.00                 & 0.49           & 0.34         \\
\textbf{}            & \textbf{MRR}     & \textbf{0.62}          & 0.04          & 0.01                 & 0.50           & 0.37         \\
\textbf{MMTweets-PT} & \textbf{MAP@1}   & \textbf{0.65}          & 0.16           & 0.00                 & 0.64           & 0.36         \\
\textbf{}            & \textbf{MAP@5}   & \textbf{0.75}          & 0.26           & 0.00                 & 0.72           & 0.48         \\
\textbf{}            & \textbf{MRR}     & \textbf{0.76}          & 0.28           & 0.01                 & 0.73           & 0.50         \\
\textbf{MMTweets-EN} & \textbf{MAP@1}   & \textbf{0.39}          & 0.01           & 0.00                 & 0.21           & 0.31         \\
\textbf{}            & \textbf{MAP@5}   & \textbf{0.52}          & 0.02           & 0.00                 & 0.30           & 0.45         \\
\textbf{}            & \textbf{MRR}     & \textbf{0.54}          & 0.02           & 0.00                 & 0.33           & 0.47         \\
\textbf{MMTweets-ES} & \textbf{MAP@1}   & \textbf{0.66}          & 0.12           & 0.00                 & 0.45           & 0.46         \\
\textbf{}            & \textbf{MAP@5}   & \textbf{0.75}          & 0.18           & 0.00                 & 0.57           & 0.53         \\
\textbf{}            & \textbf{MRR}     & \textbf{0.76}          & 0.21           & 0.01                 & 0.58           & 0.55         \\
\textbf{CLEF 21-AR}  & \textbf{MAP@1}   & \textbf{0.76}          & 0.20           & 0.07                 & 0.74           & 0.72         \\
\textbf{}            & \textbf{MAP@5}   & \textbf{0.87}          & 0.23           & 0.08                 & 0.86           & 0.84         \\
\textbf{}            & \textbf{MRR}     & \textbf{0.90}          & 0.28           & 0.11                 & 0.89           & 0.88         \\
\textbf{CLEF 22-AR}  & \textbf{MAP@1}   & 0.80          & 0.14           & 0.10                 & \textbf{0.88}           & 0.76         \\
\textbf{}            & \textbf{MAP@5}   & 0.86          & 0.17           & 0.11                 & \textbf{0.90}           & 0.80         \\
\textbf{}            & \textbf{MRR}     & 0.86          & 0.19           & 0.12                 & \textbf{0.90}           & 0.81         \\ \midrule
\textbf{Average}     & \textbf{MAP@1}   & \textbf{0.64}          & 0.11           & 0.03                 & 0.56           & 0.48         \\
\textbf{}            & \textbf{MAP@5}   & \textbf{0.73}          & 0.14           & 0.03                 & 0.64           & 0.57         \\
\textbf{}            & \textbf{MRR}     & \textbf{0.74}          & 0.17           & 0.04                 & 0.66           & 0.59         \\ \bottomrule
\end{tabular}%
}
% \caption{Results of BM25 and off-the-shelf MPT models.}
\label{tab:unxupMPT}
\end{table}

% \color{blue}
\subsection{Results of MPT Models}
\label{resUnsupmodle}

Table \ref{tab:unxupMPT} shows the results of the  of the off-the-shelf models, namely mBERT, XLM-RoBERTa, LaBSE, and USE, while also incorporating the BM25 outcomes from Table 7 (6th column - main paper) for reference. The results indicate that BM25 performs consistently well across all datasets and metrics. Among the Transformer models, LaBSE and USE deliver the most favorable outcomes across most datasets and metrics, with LaBSE exhibiting superior performance on most metrics, especially on the MMTweets-PT and CLEF 21-AR datasets. However, other Transformer models, such as mBERT and XLM-RoBERTa, tend to perform poorly in most cases.

In conclusion, LaBSE emerges as the optimal choice after BM25 as it outperforms other Transformer models in most cases. Thus, researchers and practitioners in the field of information retrieval may consider utilising LaBSE as a substitute for BM25 when they aim to achieve better performance without the need for machine translation that is required in BM25.

\begin{table}[]
\centering
\small
\caption{Evaluation results of the multistage retrieval framework using various values of $K$, which represents the number of documents re-ranked in the second stage.}
\resizebox{\columnwidth}{!}{%
\begin{tabular}{@{}lllllll@{}}
\toprule
\multicolumn{1}{c}{\textbf{Dataset}} & \multicolumn{1}{c}{\textbf{Metrics}} & \textbf{K=50}                & \textbf{K=100}               & \textbf{K=200}               & \textbf{K=300}               & \textbf{K=400}               \\ \midrule
\textbf{MMTweets-HI}                 & \textbf{MAP@1}                       & \cellcolor[HTML]{FFFFFF}0.75 & \cellcolor[HTML]{FFFFFF}0.75 & \cellcolor[HTML]{FFFFFF}0.76 & \cellcolor[HTML]{FFFFFF}0.76 & \cellcolor[HTML]{FFFFFF}0.76 \\
\textbf{}                            & \textbf{MAP@5}                       & \cellcolor[HTML]{FFFFFF}0.80 & \cellcolor[HTML]{FFFFFF}0.80 & \cellcolor[HTML]{FFFFFF}0.81 & \cellcolor[HTML]{FFFFFF}0.82 & \cellcolor[HTML]{FFFFFF}0.82 \\
\textbf{}                            & \textbf{MRR}                         & \cellcolor[HTML]{FFFFFF}0.80 & \cellcolor[HTML]{FFFFFF}0.80 & \cellcolor[HTML]{FFFFFF}0.82 & \cellcolor[HTML]{FFFFFF}0.82 & \cellcolor[HTML]{FFFFFF}0.82 \\
\textbf{MMTweets-PT}                 & \textbf{MAP@1}                       & \cellcolor[HTML]{FFFFFF}0.84 & \cellcolor[HTML]{FFFFFF}0.84 & \cellcolor[HTML]{FFFFFF}0.84 & \cellcolor[HTML]{FFFFFF}0.84 & \cellcolor[HTML]{FFFFFF}0.84 \\
\textbf{}                            & \textbf{MAP@5}                       & \cellcolor[HTML]{FFFFFF}0.89 & \cellcolor[HTML]{FFFFFF}0.89 & \cellcolor[HTML]{FFFFFF}0.89 & \cellcolor[HTML]{FFFFFF}0.89 & \cellcolor[HTML]{FFFFFF}0.89 \\
\textbf{}                            & \textbf{MRR}                         & \cellcolor[HTML]{FFFFFF}0.89 & \cellcolor[HTML]{FFFFFF}0.89 & \cellcolor[HTML]{FFFFFF}0.89 & \cellcolor[HTML]{FFFFFF}0.89 & \cellcolor[HTML]{FFFFFF}0.89 \\
\textbf{MMTweets-EN}                 & \textbf{MAP@1}                       & \cellcolor[HTML]{FFFFFF}0.45 & \cellcolor[HTML]{FFFFFF}0.44 & 0.44                         & \cellcolor[HTML]{FFFFFF}0.44 & \cellcolor[HTML]{FFFFFF}0.43 \\
\textbf{}                            & \textbf{MAP@5}                       & \cellcolor[HTML]{FFFFFF}0.59 & \cellcolor[HTML]{FFFFFF}0.58 & 0.57                         & \cellcolor[HTML]{FFFFFF}0.57 & \cellcolor[HTML]{FFFFFF}0.57 \\
\textbf{}                            & \textbf{MRR}                         & \cellcolor[HTML]{FFFFFF}0.60 & \cellcolor[HTML]{FFFFFF}0.59 & 0.58                         & \cellcolor[HTML]{FFFFFF}0.58 & \cellcolor[HTML]{FFFFFF}0.58 \\
\textbf{MMTweets-ES}                 & \textbf{MAP@1}                       & \cellcolor[HTML]{FFFFFF}0.73 & \cellcolor[HTML]{FFFFFF}0.72 & 0.73                         & \cellcolor[HTML]{FFFFFF}0.74 & \cellcolor[HTML]{FFFFFF}0.74 \\
\textbf{}                            & \textbf{MAP@5}                       & \cellcolor[HTML]{FFFFFF}0.81 & \cellcolor[HTML]{FFFFFF}0.80 & 0.80                         & \cellcolor[HTML]{FFFFFF}0.82 & \cellcolor[HTML]{FFFFFF}0.82 \\
\textbf{}                            & \textbf{MRR}                         & \cellcolor[HTML]{FFFFFF}0.81 & \cellcolor[HTML]{FFFFFF}0.80 & 0.81                         & \cellcolor[HTML]{FFFFFF}0.82 & \cellcolor[HTML]{FFFFFF}0.82 \\ \midrule
\textbf{Average}                     & \textbf{MAP@1}                       & 0.69                         & 0.69                         & 0.69                         & \textbf{0.70}                & 0.69                         \\
\textbf{}                            & \textbf{MAP@5}                       & \textbf{0.77}                & \textbf{0.77}                & \textbf{0.77}                & \textbf{0.77}                & \textbf{0.77}                \\
\textbf{}                            & \textbf{MRR}                         & \textbf{0.78}                & 0.77                         & \textbf{0.78}                & \textbf{0.78}                & \textbf{0.78}                \\ \bottomrule
\end{tabular}%
}
% \caption{Evaluation results of the multistage retrieval framework using various values of $K$, which represents the number of documents re-ranked in the second stage.}
\label{tab:docsre-rank}
\end{table}

\subsection{Influence of Number of Documents re-ranked}
\label{docsre-ranked}

Table \ref{tab:docsre-rank} shows the evaluation results of the multistage retrieval framework on MMTweets using various values of $K$, which represents the number of documents re-ranked in the second stage.
The results show that increasing the value of $K$ generally improves the model's performance, as indicated by higher MAP and MRR scores. For instance, the MAP@5 score for Hindi increases from 0.80 and 0.82, when $K$ is increased from 50 to 400, indicating a consistent improvement in the model's performance. On the other hand, for Portuguese, all three metrics remain constant. 

Overall, the results suggest that increasing the number of documents re-ranked in the second stage can improve the performance of the model, but the magnitude of the improvement may vary depending on the dataset and the evaluation metric used.
Furthermore, it's worth noting that increasing $K$ also results in a longer time taken to retrieve relevant documents, which can be a drawback in real-world applications. Therefore, in our experiments, we chose a value of $K$ as 200 to balance the trade-off between performance and efficiency.

\subsection{MPT Model in the Second Stage of Multistage Retrieval Framework}
\label{mptsecond}

Table \ref{tab:mptsecond} shows the results of using three different MPT models (mBERT, XLM-RoBERTa, and LaBSE) in the second stage of the multistage retrieval framework. Please refer to Appendix \ref{appendix:hyper} for hyperparameter details. We find that the average performance across all datasets and metrics is highest for LaBSE, followed by mBERT and then XLM-RoBERTa. In particular, LaBSE outperforms the other two models significantly in the MMTweets-PT dataset, while XLM-RoBERTa performs the worst across all datasets and metrics. Therefore, we choose LaBSE as the model for the second stage of retrieval in our experiments.

\begin{table}[]
\centering
\small
\caption{Results of different MPT models used in the second stage of the multistage retrieval framework.}
\resizebox{\columnwidth}{!}{%
\begin{tabular}{@{}llccc@{}}
\toprule
\multicolumn{1}{c}{\textbf{Dataset}} & \multicolumn{1}{c}{\textbf{Metrics}} & \textbf{mBERT} & \textbf{XLM-RoBERTa} & \textbf{LaBSE} \\ \midrule
\textbf{MMTweets-HI}                 & \textbf{MAP@1}                       & 0.49           & 0.17                 & \textbf{0.76}           \\
\textbf{}                            & \textbf{MAP@5}                       & 0.59           & 0.24                 & \textbf{0.81}           \\
\textbf{}                            & \textbf{MRR}                         & 0.60           & 0.27                 & \textbf{0.82}           \\
\textbf{MMTweets-PT}                 & \textbf{MAP@1}                       & \textbf{0.87}           & 0.14                 & 0.84           \\
\textbf{}                            & \textbf{MAP@5}                       & \textbf{0.91}           & 0.30                 & 0.89           \\
\textbf{}                            & \textbf{MRR}                         & \textbf{0.91}           & 0.34                 & 0.89           \\
\textbf{MMTweets-EN}                 & \textbf{MAP@1}                       & 0.38           & 0.03                 & \textbf{0.44}           \\
\textbf{}                            & \textbf{MAP@5}                       & 0.53           & 0.06                 & \textbf{0.57}           \\
\textbf{}                            & \textbf{MRR}                         & 0.54           & 0.11                 & \textbf{0.58}           \\
\textbf{MMTweets-ES}                 & \textbf{MAP@1}                       & 0.68           & 0.21                 & \textbf{0.73}           \\
\textbf{}                            & \textbf{MAP@5}                       & 0.78           & 0.37                 & \textbf{0.80}           \\
\textbf{}                            & \textbf{MRR}                         & 0.79           & 0.40                 & \textbf{0.81}           \\ \midrule
\textbf{Average}                     & \textbf{MAP@1}                       & 0.61           & 0.14                 & \textbf{0.69}           \\
\textbf{}                            & \textbf{MAP@5}                       & 0.70           & 0.24                 & \textbf{0.77}           \\
\textbf{}                            & \textbf{MRR}                         & 0.71           & 0.28                 & \textbf{0.78}           \\ \bottomrule
\end{tabular}%
}
% \caption{Results of different MPT models used in the second stage of the multistage retrieval framework.}
\label{tab:mptsecond}
\end{table}

% \bibliographystyle{ACM-Reference-Format}